\documentclass[nohyperref]{article}

\pdfoutput=1
\usepackage{microtype}
\usepackage{graphicx}
\usepackage{subfigure}
\usepackage{booktabs} 
\usepackage{threeparttable}

\usepackage{hyperref}

\usepackage[accepted]{icml2022}

\usepackage{amsmath}
\usepackage{amssymb}
\usepackage{mathtools}
\usepackage{amsthm}
\usepackage{xcolor}
\usepackage{colortbl}

\usepackage{color}
\usepackage{mathtools}

\usepackage{amsmath,amsfonts,bm}

\def\eqref#1{equation~\ref{#1}}

\def\1{\bm{1}}

\def\va{{\bm{a}}}

\DeclareMathAlphabet{\mathsfit}{\encodingdefault}{\sfdefault}{m}{sl}
\SetMathAlphabet{\mathsfit}{bold}{\encodingdefault}{\sfdefault}{bx}{n}

\usepackage{subfigure}
\usepackage{multirow}
\usepackage{makecell}
\usepackage{array, booktabs, arydshln}
\usepackage{amssymb}
\usepackage{graphbox}
\usepackage{url}
\usepackage{amsthm}
\usepackage{float}
\usepackage{ulem}

\allowdisplaybreaks

\newcommand{\shorte}{\textup{\texttt{=}}}

\newcommand{\name}{MATTAR}
\newcommand{\Tau}{\mathrm{T}}

\newcolumntype{L}{>{$}l<{$}}
\newcolumntype{C}{>{$}c<{$}}
\newcolumntype{R}{>{$}r<{$}}
\newcolumntype{P}[1]{>{\centering\arraybackslash}p{#1}}

\usepackage[capitalize,noabbrev]{cleveref}

\theoremstyle{plain}

\theoremstyle{definition}

\theoremstyle{remark}

\usepackage[textsize=tiny]{todonotes}

\icmltitlerunning{Multi-Agent Policy Transfer via Task Relationship Modeling}

\begin{document}

\twocolumn[
\icmltitle{Multi-Agent Policy Transfer via Task Relationship Modeling}



\icmlsetsymbol{equal}{*}

\begin{icmlauthorlist}
\icmlauthor{Rongjun Qin}{equal,nju,polixir}
\icmlauthor{Feng Chen}{equal,nju}
\icmlauthor{Tonghan Wang}{equal,thu}
\icmlauthor{Lei Yuan}{nju,polixir}
\icmlauthor{Xiaoran Wu}{thucs}
\icmlauthor{Zongzhang Zhang}{nju}
\icmlauthor{Chongjie Zhang}{thu}
\icmlauthor{Yang Yu}{nju,polixir}
\end{icmlauthorlist}

\icmlaffiliation{thu}{IIIS, Tsinghua University, Beijing, China}
\icmlaffiliation{nju}{National Key Laboratory of Novel Software Technology, Nanjing, China}
\icmlaffiliation{thucs}{Department of Computer Science and Technology, Tsinghua University, Beijing, China}
\icmlaffiliation{polixir}{Polixir Technologies, Nanjing, China}

\icmlcorrespondingauthor{Chongjie Zhang}{chongjie@tsinghua.edu.cn}
\icmlcorrespondingauthor{Zongzhang Zhang}{zzzhang@nju.edu.cn}

\icmlkeywords{Multi-agent transfer, Task relationship}

\vskip 0.3in
]



\printAffiliationsAndNotice{\icmlEqualContribution} 

\begin{abstract}
Team adaptation to new cooperative tasks is a hallmark of human intelligence, which has yet to be fully realized in learning agents. 
Previous work on multi-agent transfer learning 
accommodate teams of different sizes, heavily relying on the generalization ability of neural networks for adapting to unseen tasks. We believe that the relationship among tasks provides the key information for policy adaptation. In this paper, we try to discover and exploit common structures among tasks for more efficient transfer, and  propose to learn effect-based task representations as a common space of tasks, using an alternatively fixed training scheme. 
We demonstrate that the task representation can capture the relationship among tasks, and can generalize to unseen tasks. As a result, the proposed method can help transfer learned cooperation knowledge to new tasks after training on a few source tasks. We also find that fine-tuning the transferred policies help solve tasks that are hard to learn from scratch.
\end{abstract}

\normalem
\section{Introduction}

Cooperation in human groups is characterized by resiliency to unexpected changes and purposeful adaptation to new tasks. This flexibility and transferability of cooperation is a hallmark of human intelligence. Computationally, multi-agent reinforcement learning~\cite{tan1993multi} provides an important means for machines to imitate human cooperation. Although recent multi-agent reinforcement learning research has made prominent progress in many aspects of cooperation, such as policy decentralization~\cite{lowe2017multi, rashid2018qmix, wang2020qplex, wang2021off, cao2021linda}, communication~\cite{foerster2016learning, jiang2018learning, wang2020learning}, and organization~\cite{wang2020roma, wang2021rode, jiang2019graph}, how to realize the ability of group knowledge transfer is still an open question. 

Compared to single-agent knowledge reuse~\cite{sarltransfer}, a unique challenge faced by multi-agent transfer learning is the varying size of agent groups. The number of agents and the length of observation inputs in unseen tasks may differ from those in source tasks. To solve this problem, existing multi-agent transfer learning approaches build population-invariant~\cite{long2019evolutionary} and input-length-invariant~\cite{wang2020few} learning structures using graph neural networks~\cite{agarwal2020learning} and attentional mechanics like transformers~\cite{hu2021updet, zhou2021cooperative}. Although these methods handle varying populations and input lengths well, knowledge transfer to unseen tasks mainly depends on the inherent generalization ability of neural networks. The relationship among tasks is not fully used for more efficient transfer.

To make up for this shortage, we study the discovery and utilization of common structures in multi-agent tasks and propose Multi-Agent Transfer reinforcement learning via modelling TAsk Relationship (\name). In this learning framework, we capture the common structure of tasks by modeling the similarity among transition and reward functions of different tasks. Specifically, we train a forward model for all source tasks to predict the observation, state, and reward at the next timestep given the current observation, state, and actions. The question is how to embody the similarity as well as the difference among tasks in this forward model. We introduce difference by giving each source task a unique representation and model the similarity by generating the parameters of the forward model via a shared hypernetwork, which we call the representation explainer. 

To learn a well-formed representation space that encodes task relationship, an alternative-fixed training method is proposed to learn the task representation and representation explainer. During training, representations of source tasks are pre-defined and fixed as mutual orthogonal vectors, and the representation explainer is learned. When facing an unseen task, we fix the representation explainer and backpropagate gradients through the fixed forward model to learn the representation of the new task by a few samples.

Furthermore, we condition the agent policies on the task representation. During training, the task representation is fixed, and the policy is updated to maximize the expected return. On unseen tasks, we obtained the transferred policy by simply inserting the new task representation. The structure of the policy is also designed to be adaptable to population and observation inputs of different sizes.

We design experiments to demonstrate that the learned knowledge from three to four source tasks can be transferred to a series of unseen tasks with great success rates. We also pinpoint several advantages brought by our method other than knowledge transfer. First, fine-tuning the transferred policy on unseen tasks achieves better performance than learning from scratch, indicating that the task representation provides a good initialization point. Second, training on multiple source tasks gets better performance compared to training on them individually. This result shows that \name~also provides a method for multi-agent multi-task learning. Finally, although not designed for this goal, our structure enables better learning performance against single-task learning algorithms when trained on single tasks. 
\section{Method}

In this paper, we focus on knowledge transfer among fully cooperative multi-agent tasks that can be modelled as a Dec-POMDP~\citep{oliehoek2016concise} consisting of a tuple $G\shorte\langle I, S, A, P, R, \Omega, O, n, \gamma\rangle$, where $I$ is the finite set of $n$ agents, $s\in S$ is the true state of the environment, and $\gamma\in[0, 1)$ is the discount factor. At each timestep, each agent $i$ receives an observation $o_i\in \Omega$ drawn according to the observation function $O(s, i)$ and selects an action $a_i\in A$. Individual actions form a joint action $\va$ $\in A^n$, which leads to a next state $s'$ according to the transition function $P(s'|s, \va)$, a reward $r=R(s,\va)$ shared by all agents. Each agent has local action-observation history $\tau_i\in \Tau\equiv(\Omega\times A)^*\times \Omega$. Agents learn to collectively maximize the global action value function $Q_{tot}(s, \va)=\mathbb{E}_{s_{0:\infty}, a_{0:\infty}}[\sum_{t=0}^{\infty} \gamma^t R(s_t, \va_t) | s_0=s, \va_0=\va]$.

\begin{figure}[th]
    \centering
    \includegraphics[width=\linewidth]{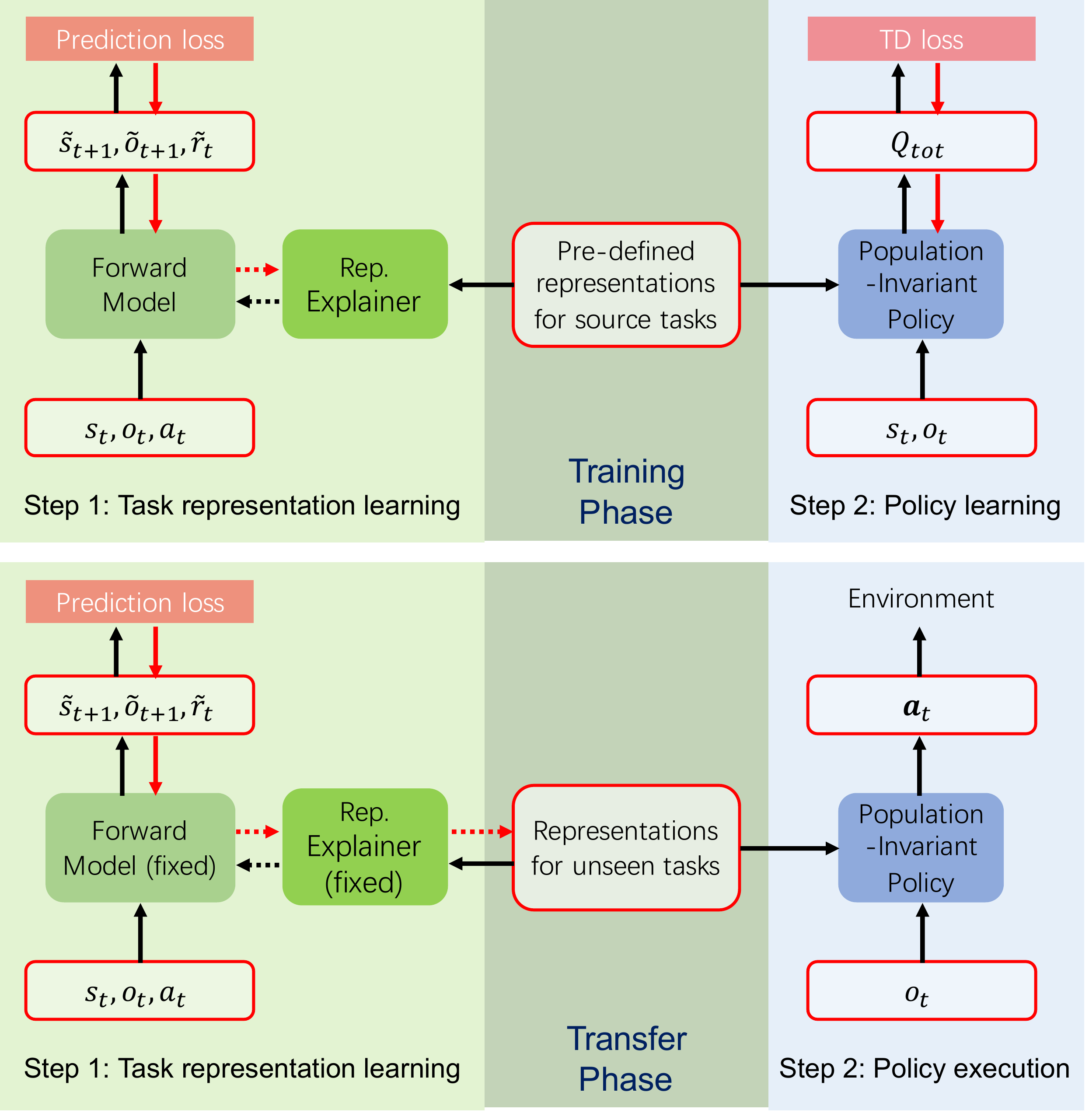}
    \vspace{-2em}
    \caption{Transfer learning scheme of our method.}
    \label{fig:overview}
\end{figure}

Overall, our framework first trains on several source tasks $\mathcal{S}=\{\mathcal{S}_i\}$ and then transfers the learned cooperative knowledge to unseen tasks $\mathcal{T}=\{\mathcal{T}_j\}$ from the same task distribution. As shown in Fig.~\ref{fig:overview}, our learning framework achieves this by designing two modules for task representation learning and task policy learning, respectively. In the following sections, we first introduce how we design the representation learning module and its learning scheme in different phases. Then, we describe the details of policy learning, including the population-invariant structure for dealing with inputs of different sizes.

\subsection{Task representation learning}\label{sec:task repre}


Our main idea in achieving knowledge transfer among multi-agent tasks is to capture and exploit both the common structure and the unique characteristics of tasks by learning task representation. A task distinguishes itself from other tasks by its transition and reward functions. Therefore, we incorporate our task representation learning component into the learning of a forward model that predicts the next state, local observations, and reward given the current state, local observations, and actions. 

We associate each task $i$ with a representation $z_i\in \mathbb{R}^m$ and expect it to reflect different properties of tasks. For modeling task similarity, all source and unseen tasks share a representation explainer, which takes as input the task representations and outputs the parameters of the forward model. The representation explainer is trained on all source tasks. Concretely, for a source task $\mathcal{S}_i$, parameters of the forward model are generated as $\eta_i=f_{\theta}(z_i)$, where $\theta$ denotes the parameters of the representation explainer. The forward model contains three predictors: $f_{\eta_i}^s$, $f_{\eta_i}^o$, and $f_{\eta_i}^r$. Given the current state $s$, agent $j$'s observation $o_j$, and action $a_j$, these predictors estimate the next state $s'$, the next observation $o_j'$, and the global reward $r$, respectively. 

A possible method for training on source tasks is to backpropagate the forward model's prediction error to update both the representation explainer and task representations. However, this training scheme leads to unexpected results in practice, which is mainly attributable to the fact that the network may ignore the task representation, and the representations may have a very small norm.

To solve this problem, we propose pre-determining the task representation for each source task and learning the representation explainer by backpropagating the prediction error. Such a method can help form an informative task representation space and build a mapping from task representation space to the space of forwarding model parameters. The question is how we pre-define these source task representations. In practice, we initialize source task representations as mutually orthogonal vectors. Specifically, we first randomly generate vectors in $\mathbb{R}^m$ for source tasks, and then use the Schimidt orthogonalization~\cite{bjorck1994numerics} on these vectors to obtain source task representations. 

With pre-defined task representations, the representation explainer is optimized to minimize the following loss function:
\begin{align}
J(\theta) := \sum_{i}^{N_{src}} J_{\mathcal{S}_i}(\theta),
\end{align}
where $N_{src}$ is the number of source tasks, and
\begin{align}
&J_{\mathcal{S}_i}(\theta) = \mathbb{E}_{\mathcal{D}}\big[\|f^s_{\eta_i}(s,o_i,a_i)-s'\|^2 \\ 
&+\lambda_1\sum_{j}\|f^o_{\eta_i}(s,o_i,a_i)-o'_i\|^2
+\lambda_2 (f_{\eta_i}^r(s, o_i, a_i) - r)^2 \big] \nonumber
\end{align}
is the per-task prediction loss of the forward model. Here, $\mathcal{D}$ is the replay buffer, and $\lambda_1, \lambda_2$ are scaling factors. 



We fix the source task representations and learn the representation explainer during the training phase. When it comes to the transfer phase, we aim to find a good task representation that can reflect the similarity of the new task to source tasks. To achieve this goal, we fix the trained representation explainer and learn the task representation by minimizing the prediction loss of the forward model on the new task. Specifically, we randomly initialize a task representation $z$, keep $\theta$ fixed, and get the forward model parameterized by $\eta=f_{\theta}(z)$. The same as the case of training the representation explainer, we compute the prediction loss for both transition and reward functions. Still, the backpropagated gradient this time is not used to update $\theta$, but to update the task representation $z$.

To keep the new task representation in the well-formed space of source task representations, we learn new task representation as a linear combination of source task representations: 
\begin{align}
z=\sum_{i=1}^{N_{src}}\mu_i z_i \ \ \text{s.t.}\ \  \mu_i \geq 0, \sum_{i=1}^{N_{src}}\mu_i=1.
\end{align}
In this way, what we are learning is the weight vector $\mu$. To make the learning more stable, we additionally optimize an entropy regularization term $\mathcal{H}(\mu)$. The final loss function for learning $z$ is:
\begin{align}
& J_{\mathcal{T}}(\mu) = \lambda\mathcal{H}(\mu) + \mathbb{E}_\mathcal{D} \big[ \|f^s_{\eta}(s,o_i,a_i)-s'\|^2  \\ 
&+\lambda_1\sum_{j}\|f^o_{\eta}(s,o_i,a_i)-o'_i\|^2 
+\lambda_2 (f_{\eta}^r(s, o_i, a_i) - r)^2 \big], \nonumber
\end{align}
where $\eta=f_{\theta}(z)$ and $z=\sum_{i=1}^{N_{src}}\mu_i z_i$.

The detailed architectures for task representation learning and forward model are described in Appendix~\ref{appx:architecture}.

\begin{figure*}[htbp]
    \centering
    \includegraphics[width=\linewidth]{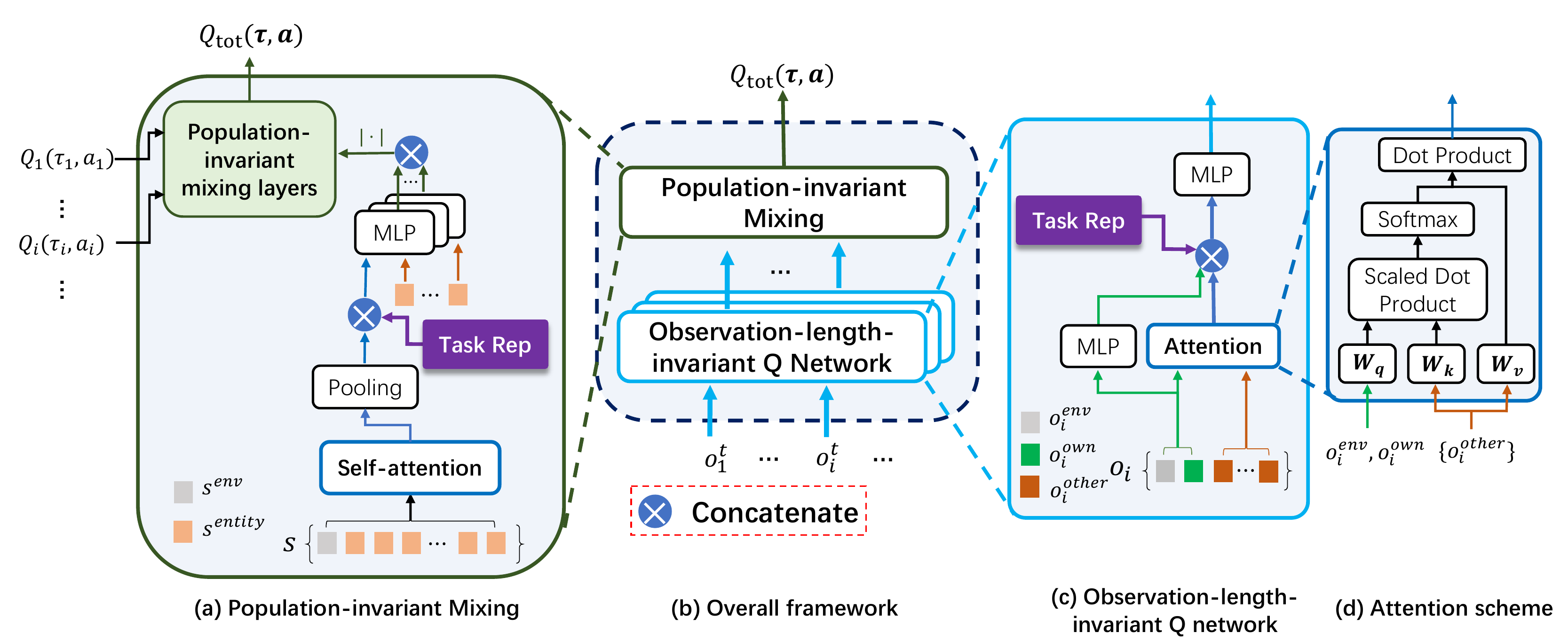}
    \vspace{-2em}
    \caption{Population-invariant network structure for policy learning.}
    \label{fig:pis}
\end{figure*}

\subsection{Task policy learning}

After the task representation is learned by modeling the transition and reward functions, it will be used to learn and transfer the policy on the source and unseen tasks.

Another difficulty faced by transferable multi-agent policy learning is that the dimension of state and observation varies across tasks with the number of agents. Thus, it is infeasible for some popular MARL algorithms like QMIX~\cite{rashid2018qmix} and MADDPG~\cite{lowe2017multi} to be directly applied to transfer knowledge by reloading the policy model since their network input sizes are fixed.

To solve this problem, we propose a \textbf{population-invariant network} structure (Fig.~\ref{fig:pis}), which can deal with the varying input sizes. This network structure is also designed to be conditioned on the learned task representations.

The population-invariant network uses the value decomposition learning framework and mainly consists of two components. Agents share an individual Q-network, and the global Q-function is learned as a combination of local Q-values. In this work, we adopt a monotonic mixing network in the style of QMIX~\cite{rashid2018qmix}, but our method is readily applied to other value decomposition methods.

In most multi-agent problems, state and observation consist of different parts corresponding to the environment information and information about other entities. We exploit this property to design the individual Q-network and the mixing network to enable them to deal with tasks with a different number of agents.

For individual Q-network, we disassemble the observation $o_i$ into the part relating to the environment $o^{env}_i$, the part relating to agent $i$ itself $o^{own}_i$, and the parts relating to other entities $\{o^{other}_j\}$. We adopt an attention mechanism where the query is generated by $o^{env}_i$ and $o^{own}_i$. This scheme learns to which entities should we pay attention:
\begin{align}
q &= \mathtt{MLP}_q([o^{env}_i, o^{own}_i]),\nonumber\\
\bm{K}&= \mathtt{MLP}_{K}([o^{other}_1,\dots,o^{other}_j,\dots]),\nonumber\\
\bm{V}&= \mathtt{MLP}_{V}([o^{other}_1,\dots,o^{other}_j,\dots]),\\
h &= \mathtt{softmax}(\frac{q\bm{K}^T}{\sqrt{d_k}})\bm{V},\nonumber
\end{align}
where $[\cdot, \cdot]$ is the vector concatenation operation, $d_k$ is the dimension of the query vector, and bold symbols are matrices. This attention mechanism helps get a fixed-dimension embedding $h$ for inputs with varying sizes. We concatenate $h$ with an embedding vector of $[o^{env}_i||e^{own}_i]$ and obtain a final fixed-dimension hidden vector. This vector is then fed into the subsequent network together with the task representation of the action value estimates. 

In some multi-agent tasks, there exist interaction actions that involve other opponents. As a result, the number of actions also varies in different tasks, presenting a challenge for knowledge transfer because conventional Q-networks have outputs with a fixed length. To handle this problem, we design a new mechanism inspired by other popular population-invariant networks~\cite{wang2019action,hu2021updet}, where we calculate Q-values for non-interaction actions and interaction actions separately. For non-interaction actions, we calculate the Q-values by directly applying a conventional deep Q-network for that the number of non-interaction actions is usually fixed. For interaction actions, we utilize a sharing network that takes as input the observation part relating to the corresponding entity, together with the concatenation of $h$ and task representation $z$. This sharing network outputs the Q-value estimation for the corresponding interaction action. We empirically compare \name~against ASN~\cite{wang2019action} in our experiments, and the detailed description of our structure for handling varying numbers of actions can be found in Appendix~\ref{appx:interaction action}.

For the mixing network, we disassemble the state $s$ into parts relating to different entities in the environment $\{s^P_j\}$. We again apply a self-attention mechanism to integrate information from these parts of the state:
\begin{align}
\bm{Q},\bm{K},\bm{V} &= \mathtt{MLP}_{Q,K,V}([s^P_1,\dots,s^P_j,\dots]),\\
\bm{H} &= \mathtt{softmax}(\frac{\bm{Q}\bm{K}^T}{\sqrt{d_k}})\bm{V},\\
h^{state} &= \mathtt{pooling}([h_1,\dots,h_j,\dots]),~h_j = \bm{H}_j. 
\end{align}
The pooling operation in the last step guarantees a fixed-dimension embedding vector. We use this embedding, together with the task representation, to generate parameters for the mixing network.

During the whole process of policy learning, we fix the task representation $z$. Compared to policy learning, which typically lasts for $2$M timesteps, the training of task representation costs few samples. In practice, we collect $50$K samples for learning task representations and the representation explainer.

When \textbf{transferring to new tasks}, we use the individual Q-network and the representation explainer trained on source tasks. We learn the task representation for $50$K timesteps and insert it into the individual Q-network. Agents execute in a decentralized manner according to this Q-network. In section~\ref{sec:exp}, we evaluate the performance of this scheme.
\section{Related Work}

\subsection{Multi-agent Transfer Learning}
Transfer learning holds the promise in improving the sample efficiency of reinforcement learning~\cite{sarltransfer} and multi-agent reinforcement learning~\cite{silva2021transfer}. The basic idea behind multi-agent transfer learning is reusing knowledge from other tasks or other learning agents, which corresponds to inter-agent transfer and intra-agent transfer, respectively. It is expected that this knowledge reuse can accelerate coordination compared to learning from scratch. 

The intra-agent transfer paradigm aims at investigating how to best reuse knowledge from other agents~\cite{,yang2021efficient} with different sensors or (possibly) internal representations via communication. DVM~\cite{DBLP:conf/iros/WadhwaniaKOH19} treats the multi-agent problem as a multi-task problem to combine knowledge from multiple tasks, and then distills 
this knowledge by a value matching mechanism. LeCTR~\cite{DBLP:conf/aaai/OmidshafieiKLTR19} learns to teach in a multi-agent environment. \citeauthor{yang2021efficient}~\citeyearpar{yang2021efficient} takes a further step by proposing an option-based policy transfer for multi-agent cooperation. 

On the other hand, inter-agent transfer refers to reusing knowledge from previous tasks, focusing on transferring knowledge across multi-agent tasks. The varying populations and input lengths impede the transfer among agents, with which the graph neural networks~\cite{wang2020few} and transformer mechanism~\cite{zhou2021cooperative} play promising roles. DyMA-CL~\cite{wang2020few} designs various transfer mechanisms across curricula to accelerate the learning process based on a dynamic agent-number network. EPC~\cite{long2019evolutionary} proposes a curriculum learning paradigm via an evolutionary approach to scale up the population number of agents. UPDeT~\cite{hu2021updet} and PIT~\cite{zhou2021cooperative} make full use of the generalization ability of the transformer to accomplish the multi-agent cooperation and transfer between tasks. Although these methods can accurately the learning efficiency for multi-agent reinforcement learning somehow, they neglect task representation in multi-agent tasks, resulting in low transfer efficiency in complex scenarios.

\subsection{Multi-agent Representation Learning}
Learning effective representation in MARL is receiving significant attention for its effectiveness in solving many important problem. CQ-Learning~\cite{staterepresentation} learns to adapt the state representation for multi-agent systems to coordinate with other agents. \citeauthor{GroverAGBE18}~\citeyearpar{GroverAGBE18} learns useful policy representations to model agents behavior in a multi-agent system. LILI \cite{Latentopponent} learns latent representations to capture the relationship between its behavior and the other agent's future strategy and uses it to influence the other agent. RODE~\cite{wang2021rode} uses an action encoder to learn action representations and applies clustering methods to decompose joint action spaces into restricted role action spaces to reduce the policy search space for multi-agent cooperation. MAR~\cite{zhang2021learning} learns meta representation for multi-agent generalization.
Unlike previous work, our approach focuses on learning meaningful task representation for efficient transfer learning in multi-agent reinforcement learning. 
\section{Experiments}\label{sec:exp}


In this section, we design experiments to evaluate the following properties of the proposed method. (1) Generalizability to unseen tasks. Can our learning framework extract knowledge from multiple source tasks and transfer the cooperation knowledge to unseen tasks? Do task representations play an indispensable role in transfer (see Sec.~\ref{sec:exp-generalize})?  (2) A good initialization for policy fine-tuning. Fine-tuning the transferred policy can help us succeed in super hard tasks, which can not be solved effectively when learning from scratch (see Sec.~\ref{sec:exp-finetune}).  (3) Benefits of multi-task training. Our multi-task learning paradigm helps the model better leverage knowledge of different source tasks to boost the learning performance compared to training on source tasks individually (see Sec.~\ref{sec:exp-multitask}). (4) Performance advantages on single tasks. Although we did not design our framework for better performance on single-task training. We find our network performs better against the underlying algorithms (see Sec.~\ref{sec:exp-bonus}).


We evaluate \name~on the benchmark of SMAC~\cite{samvelyan2019starcraft} based on PyMARL\footnote{Our experiments are all based on the PyMARL framework, which uses SC2.4.6.2.6923. Note that performance is not always comparable among versions.}. To better fit the multi-task training setting, we extend the original SMAC maps and sort out three task series. The first series consists of tasks with varying numbers of Marines. The second series involves ally and enemy teams of Stalkers and Zealots. The third series contains tasks with teams consisting of Marines, Maneuvers, and Medivacs. The detailed description of these tasks is described in Appendix~\ref{appx:smac}. To test the transferability of \name, we divide the tasks in each series into source tasks and unseen tasks.

To ensure fair evaluation, we carry out all the experiments with five random seeds, and the results are shown with a 95\% confidence interval. For a more comprehensive description of experimental details, please refer to Appendix~\ref{appx:experimental details}.

\subsection{Generalizability to unseen tasks}\label{sec:exp-generalize}
As the major desiderata of our method, we expect that \name~has better transfer performance on unseen tasks. We compare our method against the state-of-the-art multi-agent transfer method UPDeT~\cite{hu2021updet} which is based on the transformer. 

UPDeT transfers knowledge from a single source task. For a fair comparison, we transfer from each source task to every unseen task and calculate the best (UPDeT-b) and mean (UPDeT-m) performance on each unseen task. For the test phase, we conduct transfer experiments on both source tasks and unseen tasks. We carry out experiments on all the three series of tasks, and record results in Tables~\ref{table:sz test}$\sim$\ref{table:m test}.

We find that UPDeT-b outperforms UPDeT-m in all cases, indicating that the similarity between source task and target task significantly influences the transfer performance, and a good source task selection is important for this kind of transfer algorithm. In contrast, our algorithm obtains a general model without requiring much prior knowledge about task relationships. Moreover, \name~shows better transfer performance on unseen tasks than UPDeT-b, especially in complex settings requiring sophisticated coordination like the MMM series.

To investigate the role of task representations in our method, we ablate this component by inserting a zero vector of the same dimension as our task representation into the policy network. We can see that this ablation ($\mathtt{w/o\ task\ rep.}$) dramatically underperforms \name~on most unseen tasks. For each example, after trained on $\mathtt{2s3z}$, $\mathtt{3s5z}$, and $\mathtt{3s5z\_vs\_3s6z}$, \name~achieves a win rate of $0.99$ on $\mathtt{3s4z}$ but a win rate of $0.20$ without the help of task representation. We thus conclude that \textbf{task representations play an indispensable role} in policy transfer.
\begin{table}[t!] 
  \caption{Transfer performance on a series of SMAC maps involving Stalkers and Zealots.}\label{table:sz test}
  \vspace{0.2em}
  \centering
    \begin{tabular}{lcccc}
    \toprule
    {} &  \name & \makecell{w/o \\ task rep.} &   UPDeT-b &   UPDeT-m \\
    \toprule
    \multicolumn{5}{c}{Source Tasks} \\
    \cmidrule(lr){1-1}
    \cmidrule(lr){2-2}
    \cmidrule(lr){3-3}
    \cmidrule(lr){4-4}
    \cmidrule(lr){5-5}
    2s3z         &      \textbf{1.0} &      0.15 &    0.97 &    0.67 \\
    3s5z         &      \textbf{1.0} &      0.14 &    0.91 &    0.52 \\
    35\_36 &     \textbf{0.53} &       0.0 &    0.24 &    0.08 \\
    \toprule
    \multicolumn{5}{c}{Unseen Tasks} \\
    \cmidrule(lr){1-1}
    \cmidrule(lr){2-2}
    \cmidrule(lr){3-3}
    \cmidrule(lr){4-4}
    \cmidrule(lr){5-5}
    23\_24 &     \textbf{0.02} &       0.0 &     0.0 &     0.0 \\
    3s4z         &     \textbf{0.99} &       0.2 &    0.98 &    0.57 \\
    4s7z         &     \textbf{0.73} &      0.01 &    0.29 &    0.14 \\
    47\_48 &     \textbf{0.14} &       0.0 &     0.0 &     0.0 \\
    \bottomrule
    \end{tabular}
    \begin{tablenotes}
        \item Note: 35\_36 is short for 3s5z\_vs\_3s6z, 23\_24 is short for 2s3z\_vs\_2s4z, and 47\_48 is short for 4s7z\_vs\_4s8z.
    \end{tablenotes}
  \vspace{-1em}
\end{table}
\begin{table}[t!]
  \caption{Transfer performance on a series of SMAC maps involving Marines, Maneuvers, and Medivacs.}\label{table:MMM test}
  \vspace{0.2em}
  \centering
    \begin{tabular}{lcccc}
    \toprule
    {} &  \name & \makecell{w/o \\ task rep.} &   UPDeT-b &   UPDeT-m \\
    \toprule
    \multicolumn{5}{c}{Source Tasks} \\
    \cmidrule(lr){1-1}
    \cmidrule(lr){2-2}
    \cmidrule(lr){3-3}
    \cmidrule(lr){4-4}
    \cmidrule(lr){5-5}
    MMM  &      \textbf{1.0} &      0.36 &     \textbf{1.0} &    0.96 \\
    MMM2 &     \textbf{0.92} &      0.14 &    0.78 &    0.39 \\
    MMM4 &     \textbf{0.93} &      0.07 &    0.76 &    0.36 \\
    \toprule
    \multicolumn{5}{c}{Unseen Tasks} \\
    \cmidrule(lr){1-1}
    \cmidrule(lr){2-2}
    \cmidrule(lr){3-3}
    \cmidrule(lr){4-4}
    \cmidrule(lr){5-5}
    MMM0 &      \textbf{1.0} &      0.39 &    0.43 &     0.4 \\
    MMM1 &     \textbf{0.99} &      0.04 &    0.77 &    0.51 \\
    MMM3 &     \textbf{0.86} &      0.01 &     0.7 &    0.35 \\
    MMM5 &     \textbf{0.24} &       0.0 &     0.0 &     0.0 \\
    MMM6 &     \textbf{0.02} &       0.0 &     0.0 &     0.0 \\
    \bottomrule
    \end{tabular}
\end{table}
\begin{table}[t]
  \caption{Transfer performance on a series of SMAC maps involving Marines.}\label{table:m test}
  \vspace{0.2em}
  \centering
    \begin{tabular}{lcccc}
    \toprule
    {} &  \name & \makecell{w/o \\ task rep.} &   UPDeT-b &   UPDeT-m \\
    \toprule
    \multicolumn{5}{c}{Source Tasks} \\
    \cmidrule(lr){1-1}
    \cmidrule(lr){2-2}
    \cmidrule(lr){3-3}
    \cmidrule(lr){4-4}
    \cmidrule(lr){5-5}
    5m         &      \textbf{1.0} &      0.94 &     \textbf{1.0} &    0.82 \\
    5m\_6m   &     0.69 &      0.08 &    \textbf{0.93} &    0.24 \\
    8m\_9m   &     \textbf{0.96} &      0.67 &    0.84 &     0.4 \\
    10m\_11m &     0.81 &      0.62 &    \textbf{0.92} &    0.48 \\
    \toprule
    \multicolumn{5}{c}{Unseen Tasks} \\
    \cmidrule(lr){1-1}
    \cmidrule(lr){2-2}
    \cmidrule(lr){3-3}
    \cmidrule(lr){4-4}
    \cmidrule(lr){5-5}
    3m         &     \textbf{0.94} &      0.72 &    0.42 &    0.17 \\
    4m         &     \textbf{0.99} &      0.94 &    0.97 &    0.57 \\
    4m\_5m         &     \textbf{0.02} &       0.0 &     0.0 &     0.0 \\
    6m         &      \textbf{1.0} &       \textbf{1.0} &     \textbf{1.0} &    0.98 \\
    6m\_7m   &     0.76 &       0.4 &    \textbf{0.82} &    0.33 \\
    7m         &      \textbf{1.0} &       \textbf{1.0} &     \textbf{1.0} &    0.97 \\
    7m\_8m   &     \textbf{0.83} &      0.78 &     0.7 &    0.36 \\
    8m         &      \textbf{1.0} &       \textbf{1.0} &     \textbf{1.0} &    0.85 \\
    9m         &      \textbf{1.0} &      0.99 &     \textbf{1.0} &     0.8 \\
    9m\_10m  &     0.84 &      0.77 &    \textbf{0.92} &    0.46 \\
    10m        &      \textbf{1.0} &       \textbf{1.0} &     \textbf{1.0} &    0.72 \\
    10m\_12m &     \textbf{0.08} &      0.01 &    0.07 &    0.02 \\
    \bottomrule
\end{tabular}
    \begin{tablenotes}
        \item Note: x\_y is short for x\_vs\_y, e.g. 5m\_6m is short for 5m\_vs\_6m.
    \end{tablenotes}
    \vspace{-1em}
\end{table}

\subsection{A good initialization for policy fine-tuning}\label{sec:exp-finetune}

When evaluating the performance of our method on unseen tasks, we only train the task representations but remain other parts of our framework unchanged. In this section, we investigate the performance of \name~after fine-tuning. 

\begin{figure*}[t!]
\centering
\subfigure[10m\_vs\_12m]{
\begin{minipage}[t]{0.32\linewidth}
\centering
\includegraphics[width=\linewidth]{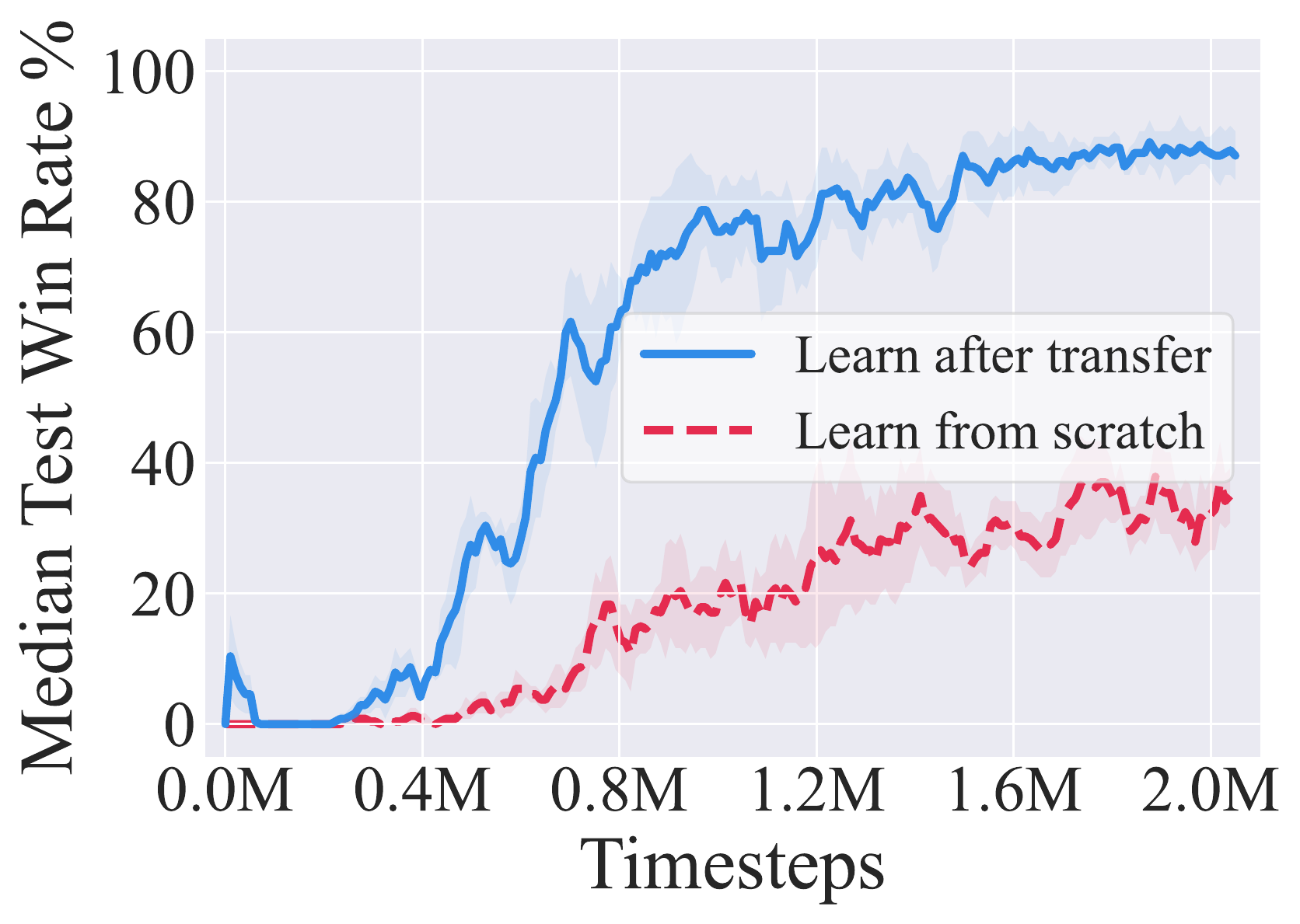}
\end{minipage}%
}%
\subfigure[3s5z\_vs\_3s7z]{
\begin{minipage}[t]{0.32\linewidth}
\centering
\includegraphics[width=\linewidth]{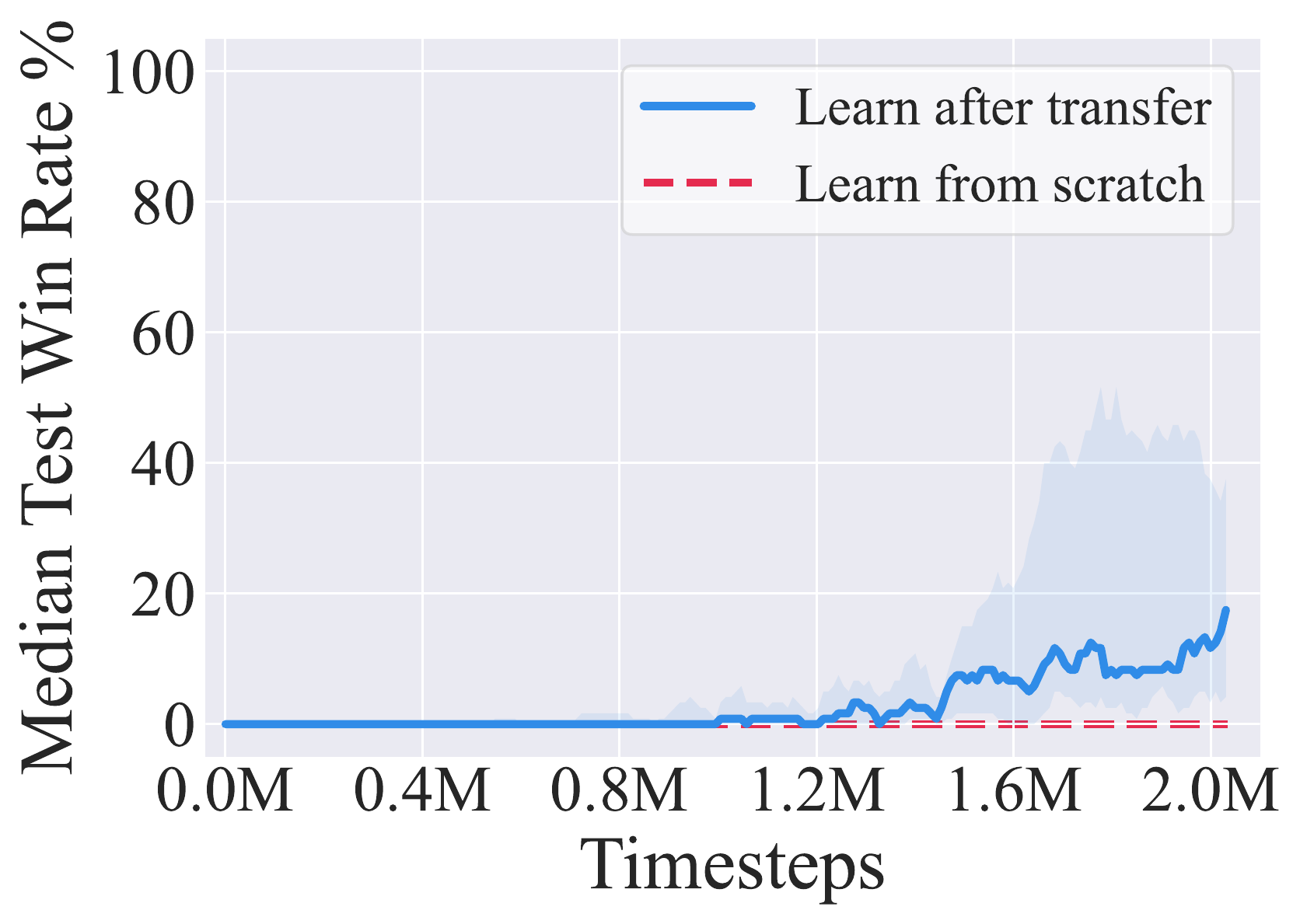}
\end{minipage}%
}%
\subfigure[MMM6]{
\begin{minipage}[t]{0.32\linewidth}
\centering
\includegraphics[width=\linewidth]{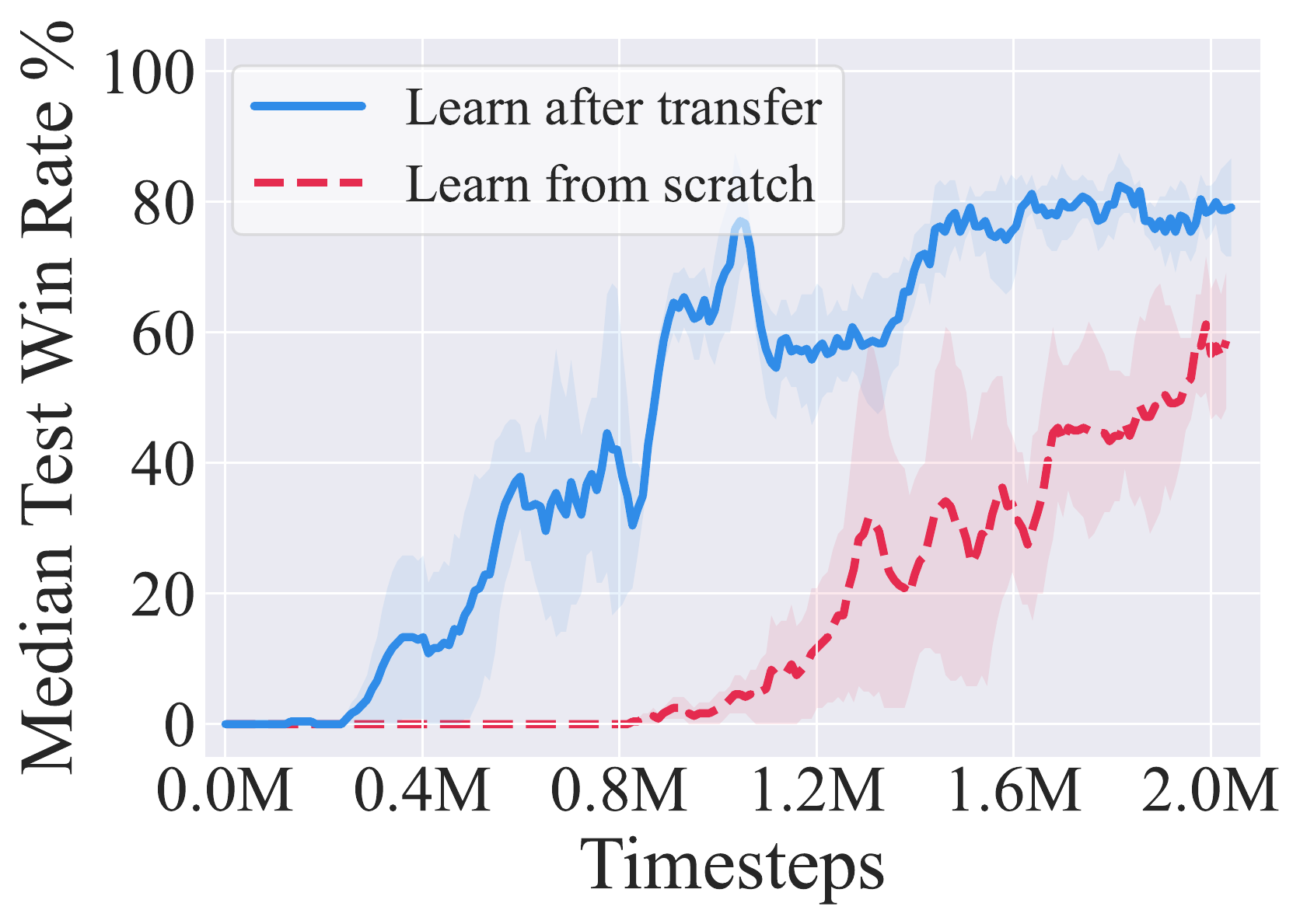}
\end{minipage}
}%
\centering
\caption{On unseen tasks: task representations provide a good initialization. Fine-tuning the policy can effectively learn cooperation policies on tasks which cannot be solved effectively when learning from scratch. }
\label{fig:exp4}
\end{figure*}

Specifically, we train the task representations for an unseen task for $50$K timesteps, then we fix it and train the policy network for $2$M timesteps. In Fig.~\ref{fig:exp4}, we compare this performance against learning from scratch on three unseen tasks from three different task series. 

We observe that the task representation provides a good initialization. For example, on $\mathtt{10m\_vs\_12m}$, after $2$M training samples, \name~with task representation converges to the win rate of around 0.86, while training solely on this task can only achieve a win rate of about $0.4$. Furthermore, with the help of task representation, \name~solves $\mathtt{3s5z\_vs\_3s7z}$, a task harder than the super hard $\mathtt{3s5z\_vs\_3s6z}$, which cannot be solved when learning from scratch. 

On other tasks where learning from scratch can obtain a satisfactory win rate, inserting the task representation can also improve the sample efficiency. For example, on $\mathtt{MMM6}$, after experiencing $0.8$M training samples, \name~wins in around $35\%$ of the games, compared to the zero win rate obtained by learning from scratch.

\subsection{Benefits of multi-task learning}\label{sec:exp-multitask}

\begin{figure*}[htbp]
\subfigure[5m\_vs\_6m]{\includegraphics[width=0.32\linewidth]{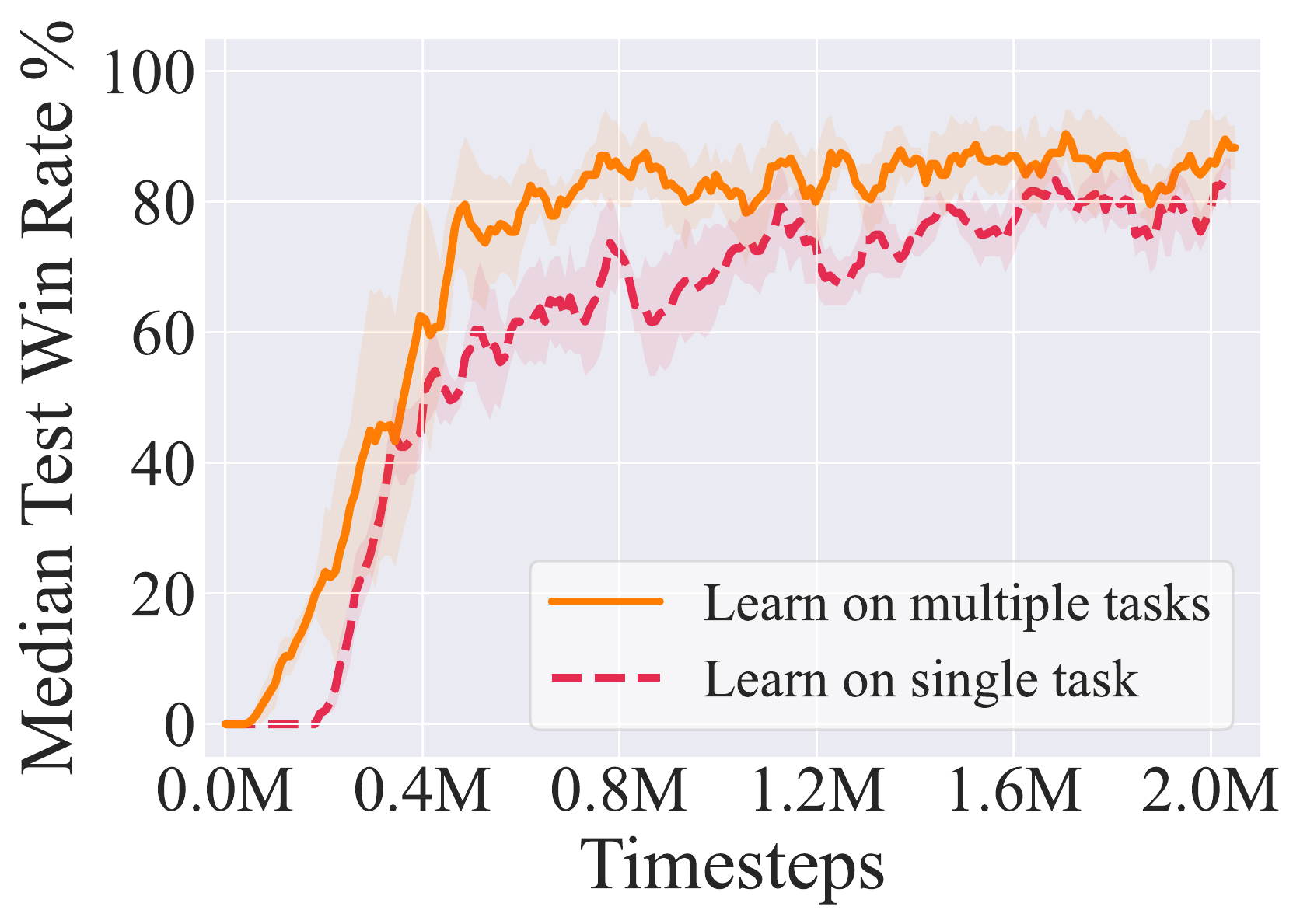}}
\subfigure[3s5z\_vs\_3s6z]{
\includegraphics[width=0.32\linewidth]{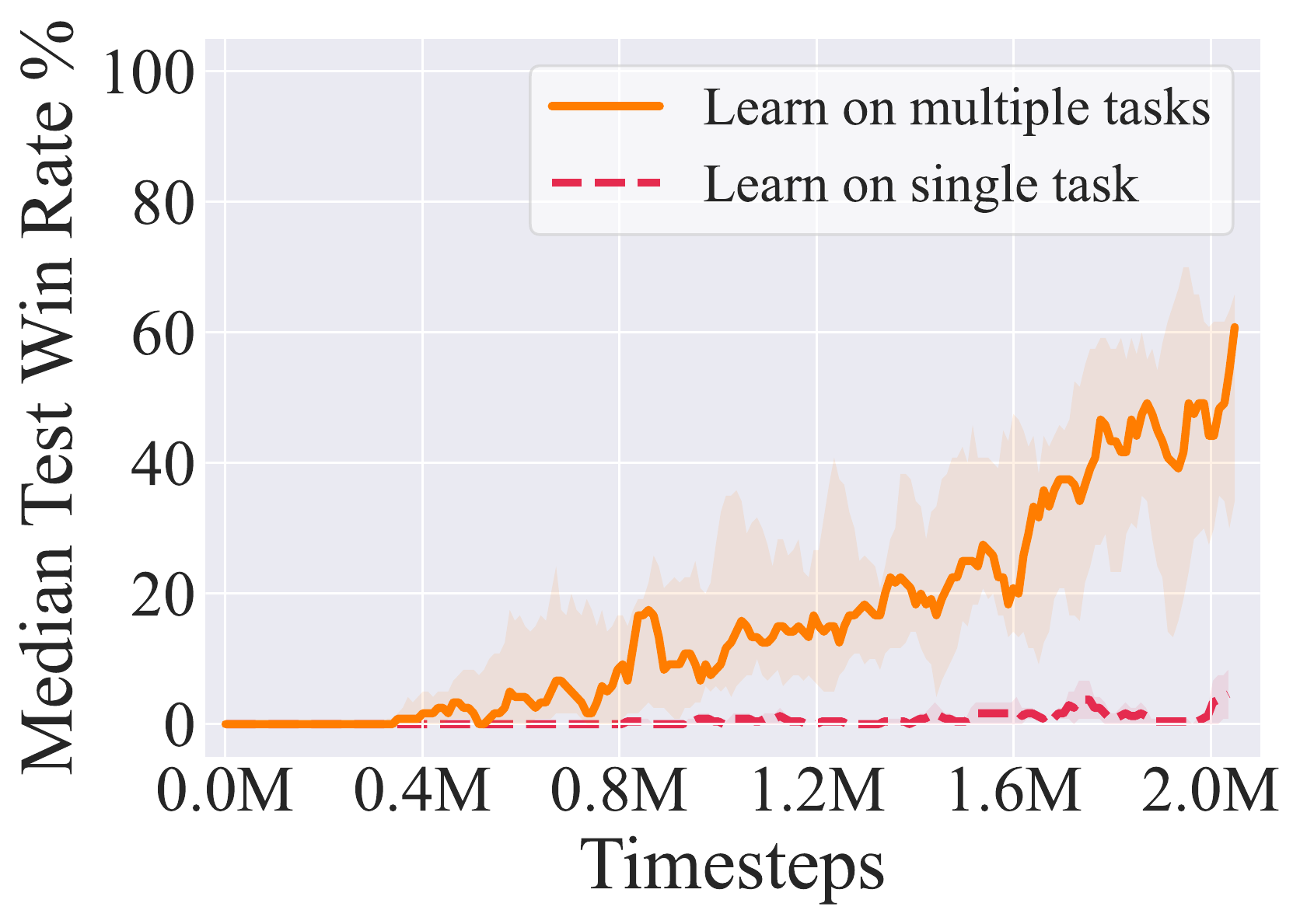}}%
\subfigure[MMM2]{
\includegraphics[width=0.32\linewidth]{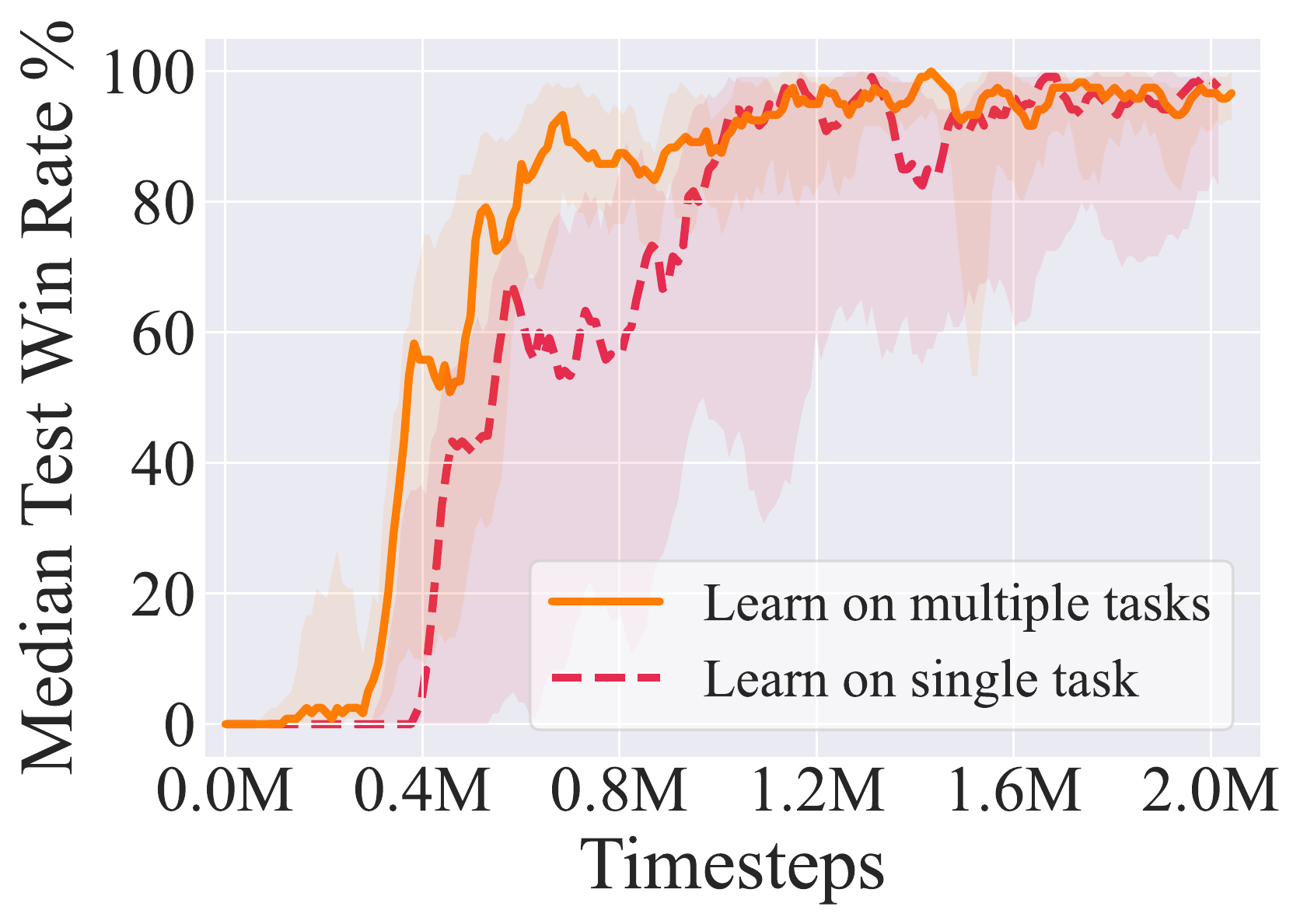}}%
\caption{On source tasks: \name~provides a framework for multi-agent multi-task learning. Training on multiple tasks helps improve performance than learning on a single task.}
\label{fig:exp3}
\end{figure*}

\begin{table}[h]
  \caption{Task representations for unseen tasks are learned as a linear combination of source tasks' representations. This table shows the learned coefficients of the linear combination. \emph{Unseen tasks mainly leverage knowledge from the most similar source task.}}\label{table:weight}
  \vspace{0.2em}
  \centering
    \begin{tabular}{ccccc}
    \toprule
    {Unseen} & \multicolumn{4}{c}{Source Tasks} \\
    \toprule
    {} & 5m & 5m\_6m & 8m\_9m & 10m\_11m \\

    \cmidrule(lr){1-1}
    \cmidrule(lr){2-2}
    \cmidrule(lr){3-3}
    \cmidrule(lr){4-4}
    \cmidrule(lr){5-5}
    4m & \textbf{0.61} & 0.13 & 0.14 & 0.12 \\
    10m\_12m & \textbf{0.43} & 0.07 & 0.08 & \textbf{0.43} \\
    \midrule
    {} & 2s3z & 3s5z & 3s5z\_3s6z & {} \\
    
    \cmidrule(lr){1-1}
    \cmidrule(lr){2-2}
    \cmidrule(lr){3-3}
    \cmidrule(lr){4-4}
    \cmidrule(lr){5-5}
    3s4z & 0.21 & \textbf{0.59} & 0.21 & {}\\
    3s5z\_3s7z & 0.18 & 0.21 & \textbf{0.61} & {}\\
    \midrule
    {} & MMM & MMM2 & MMM4 & {} \\
    \cmidrule(lr){1-1}
    \cmidrule(lr){2-2}
    \cmidrule(lr){3-3}
    \cmidrule(lr){4-4}
    \cmidrule(lr){5-5}
    MMM0 & \textbf{0.44} & 0.15 & 0.40 & {}\\
    MMM6 & 0.30 & 0.14 & \textbf{0.56} & {}\\
    \bottomrule
    \end{tabular}
\end{table}

During training, \name~adopts a scheme where multiple sources are learned simultaneously. Our aim is to leverage knowledge from more tasks and to be able to generalize the learned knowledge to a larger set of unseen tasks. Empirically, we find that this multi-task training setting helps not only unseen tasks but also the source tasks themselves.

In Fig.~\ref{fig:exp3}, we present the performance of \name~on source tasks when training with multiple tasks and a single task. The experiments are carried out three tasks from three different series. We can see that training on multiple tasks significantly boosts learning performance. For example, on $\mathtt{3s5z\_vs\_3s6z}$, after $2$M training samples, \name~with multiple tasks converges to the win rate of around $0.6$, while training solely on this task can only achieve a win rate of about $0.05$.

These results demonstrate that \emph{\name~also provides a good learning framework for multi-agent multi-task learning}. It can leverage experience on other tasks to improve performance on a similar task.

\subsection{Bonus: performance on single-task training}\label{sec:exp-bonus}

Although not designed for this goal, we find that \name~can outperform state-of-the-art MARL algorithms when trained on some single tasks. Specifically, we use random task representations and train \name~from scratch. We compare our method with two state-of-the-art value-based MARL baselines (QMIX~\cite{rashid2018qmix} and QPLEX~\cite{wang2020qplex}), a role-based learning algorithm (RODE)~\cite{wang2021rode}, and the underlying algorithm of \name~which considers the Q values of interaction actions separately (ASN)~\cite{wang2019action}.

\begin{figure*}[htbp]
\centering
\subfigure[3s5z]{\includegraphics[width=0.32\linewidth]{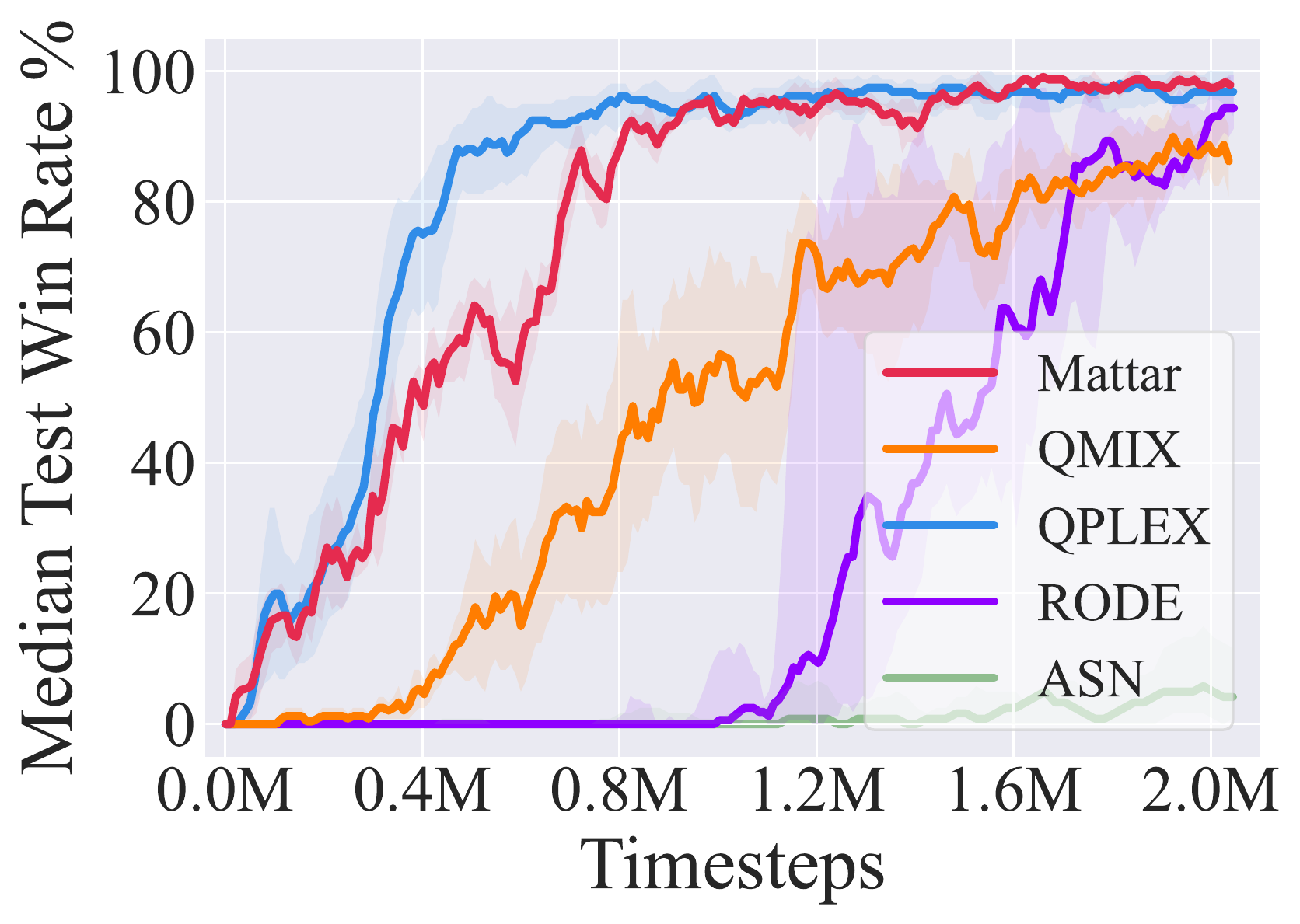}}%
\subfigure[5m\_vs\_6m]{\includegraphics[width=0.32\linewidth]{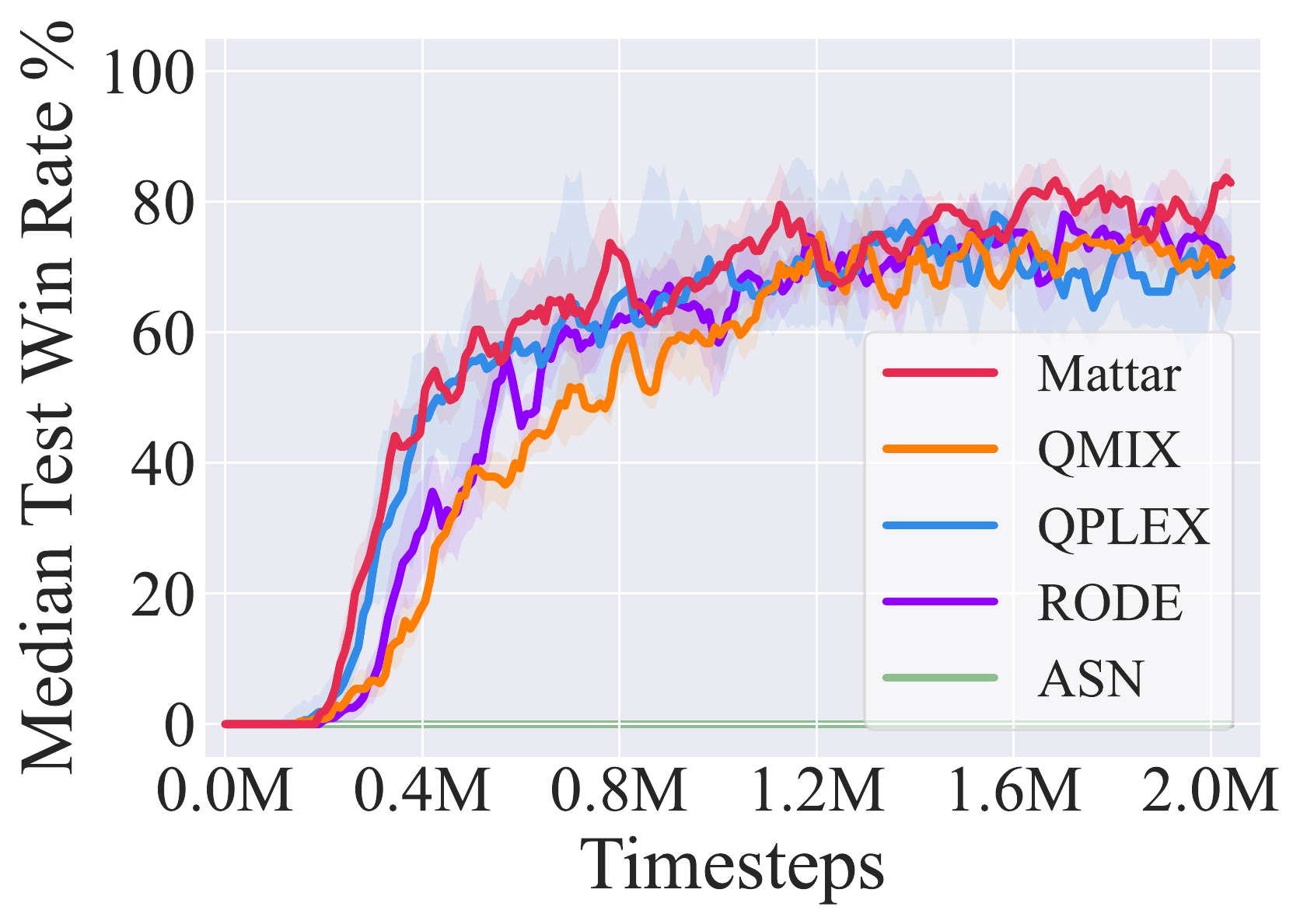}}%
\subfigure[MMM2]{\includegraphics[width=0.32\linewidth]{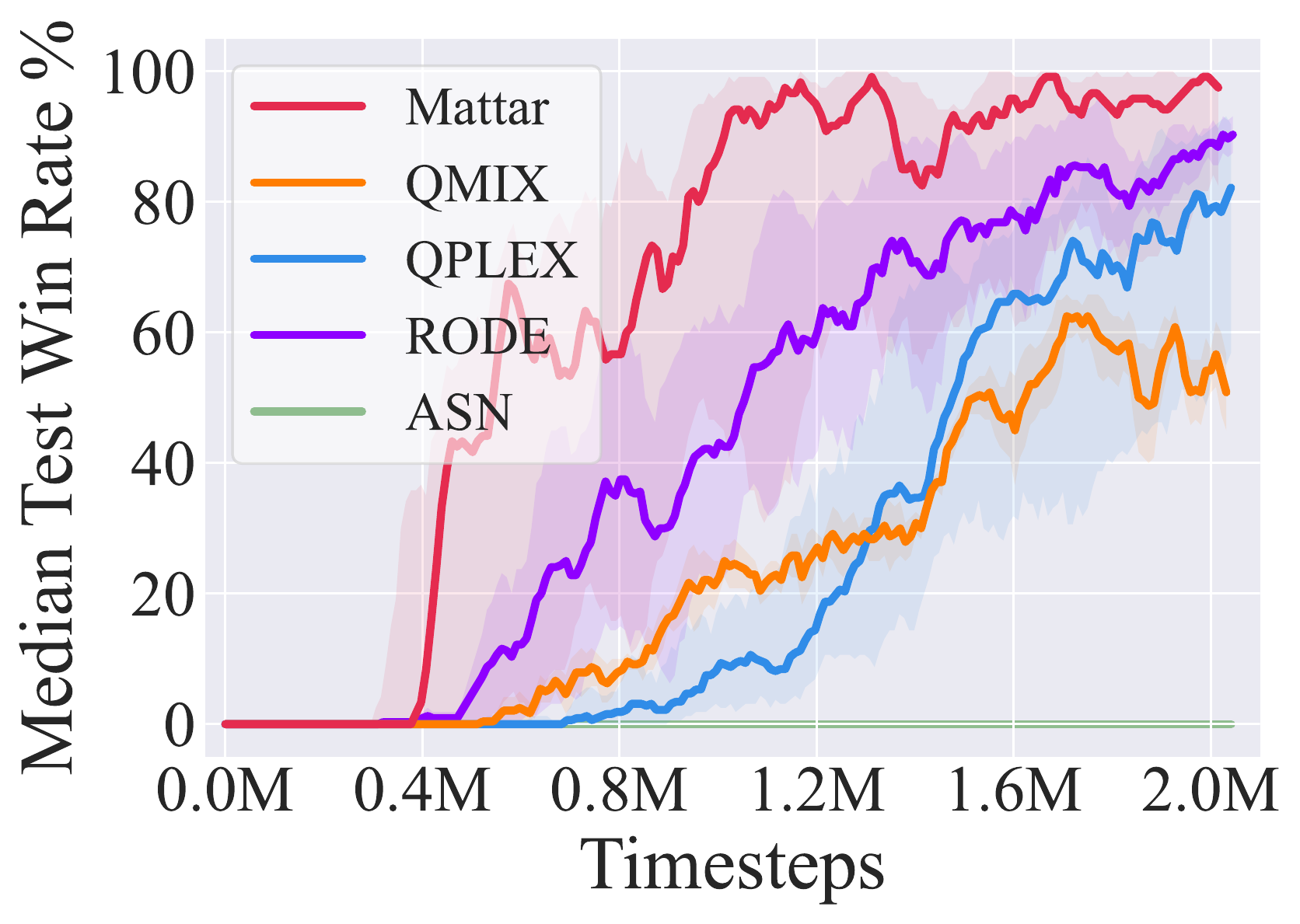}}%
\centering
\caption{On single tasks: when learning from scratch on single tasks, \name~exhibits superior performance. For performance on the whole SMAC benchmark, please refer to Appendix~\ref{appx:full performance}.}
\label{fig:exp2}
\end{figure*}

Figure~\ref{fig:exp2} shows the learning curves of different methods. We find that our population-invariant network structure achieves comparable performance in all tasks. It is worth noting that this structure even significantly outperforms other algorithms on the super hard map $\mathtt{MMM2}$.

In Appendix~\ref{appx:full performance}, we show the performance of \name~on more SMAC maps. It can be observed there that \name~also has comparable performance against baseline algorithms on these maps. Given that our underlying algorithm is QMIX, this is an inspiring result. We hypothesize that this result is because our self-attention scheme increases the representational capacity by learning to attend to appropriate entities in the environment.

\subsection{Learned linear combination of the representations for unseen tasks}

When encountering an unseen task, we first learn its representation as a linear combination of the representations for all source tasks. Specifically, we directly update the coefficients of this linear combination by backpropagating the prediction error of the forward model. For a deeper understanding of how our method transfers the learned knowledge, we are curious about the learned coefficients of the linear combination because they contain much information about the relationship between source and unseen tasks.

For each series of tasks, we show the coefficients of two unseen tasks in Table~\ref{table:weight}. We observe that the largest coefficient typically corresponds to the source task, which is the most similar to the unseen task. In the first series, $\mathtt{5m}$ is the closest source task to the unseen task $\mathtt{4m}$, and the coefficient of $\mathtt{5m}$ takes up $61\%$ of all the coefficients. In the second series, $\mathtt{3s5z\_vs\_3s6z}$ is the closest source task to the unseen task $\mathtt{3s5z\_vs\_3s7z}$. Correspondingly, the coefficient of $\mathtt{3s5z\_vs\_3s6z}$ also takes up $61\%$ of all the coefficients, indicating an important role of this source task. A similar phenomenon can be observed in the case of the third series and in $\mathtt{MMM}$/$\mathtt{MMM0}$ and $\mathtt{MMM4}$/$\mathtt{MMM6}$.

There are also some exceptions. For example, in the unseen task $\mathtt{10m\_vs\_12m}$, the coefficients of two source tasks, $\mathtt{5m}$ and $\mathtt{10m\_vs\_11m}$, are equal, and they together take up $86\%$ of all the coefficients. While $\mathtt{10m\_vs\_11m}$ is very similar to $\mathtt{10m\_vs\_12m}$, the policy for solving $\mathtt{5m}$ is very different from that for $\mathtt{10m\_vs\_12m}$. However, in $\mathtt{10m\_vs\_12m}$, agents usually first form some groups to set up an attack curve quickly. Therefore, on a local battlefield, there are around five allies fighting against a similar number of enemies. In this case, the policy learned from $\mathtt{5m}$ can be used locally. We hypothesize that \name~can learn a mixing of source task policies to solve a new task.

We conclude that, in our learning framework, unseen tasks can effectively leverage cooperation knowledge from the most similar source tasks and occasionally use knowledge from mixing of source tasks.
\section{Closing Remarks}

In this paper, we study the problem of cooperative multi-agent transfer reinforcement learning. Previous work on multi-agent transfer mainly deals with the varying population and input lengths, relying on the generalization ability of neural networks for cooperation knowledge transfer, ignoring the task relationship. Our method improves the transfer performance by learning task representations that capture the difference and similarities among tasks. When facing a new task, our method only needs to obtain a new representation before transferring the learned knowledge to it. Taking advantage of task relationship mining, \name~achieves the best transfer performance and exhibits some other advantages of this algorithm.

An important direction in the future is the transfer among tasks from different task distributions. This paper does not investigate whether the learned cooperation policies can be transferred to very dissimilar tasks. Another interesting problem is whether a linear combination of source tasks' representations can fully represent unseen tasks.


\bibliography{example_paper}
\bibliographystyle{icml2022}

\newpage
\ULforem
\appendix
\onecolumn
\section{Appendix}
\subsection{StarCraft II Micromanagement Benchmark (SMAC)}\label{appx:smac}
SMAC~\cite{samvelyan2019starcraft} is a combat scenario of StarCraft II unit micromanagement tasks, which is a popular benchmark for multi-agent reinforcement learning algorithms. 
We consider a partial observation setting, where an agent can only see a circular area around it with a radius equal to the sight range, which is default to $9$. We train the ally units with reinforcement learning algorithms to beat enemy units controlled by the built-in AI. At the beginning of each episode, allies and enemies are generated at specific regions on the map. Every agent takes action from the discrete action space at each timestep, including the following actions: no-op, move [direction], attack [enemy id], and stop. Under the control of these actions, agents can move and attack in continuous maps. Agents will get a shard reward equal to the total damage done to enemy units at each timestep. Killing each enemy unit and winning the combat (killing all the enemies) will bring additional bonuses of $10$ and $200$, respectively. We consider three settings which contain multiple micromanagement maps, and each includes various single tasks, the detailed descriptions are shown in Tables~\ref{SMACMMM}$\sim$\ref{SMACm}.

\begin{table*}[h]
\vspace{-1em}
  \caption{Description of MMM\_series SMAC maps. }
  \label{SMACMMM}
  \centering
  \newcommand{\tabincell}[2]{\begin{tabular}{@{}#1@{}}#2\end{tabular}}
  \begin{tabular}{|c|c|c|c|c|}
    \hline
    Map Name & Ally Units & Enemy Units & Type & Difficulty\\
     \hline
    MMM0 & \tabincell{l}{1 Medivac, \\2 Marauders, \\5 Marines} & \tabincell{l}{1 Medivac, \\2 Marauders, \\5 Marines} & Asymmetric \& Heterogeneous & Easy\\
    \hline
      {MMM} & \tabincell{l}{1 Medivac, \\2 Marauders, \\7 Marines} & \tabincell{l}{1 Medivac, \\2 Marauders, \\7 Marines} & Asymmetric \& Heterogeneous & Easy\\
    \hline
    MMM1 & \tabincell{l}{1 Medivac, \\1 Marauders, \\7 Marines} & \tabincell{l}{1 Medivac, \\2 Marauders, \\7 Marines} & Asymmetric \& Heterogeneous & Hard\\
    \hline
    MMM2 & \tabincell{l}{1 Medivac, \\2 Marauders, \\7 Marines} & \tabincell{l}{1 Medivac, \\3 Marauders, \\8 Marines} & Asymmetric \& Heterogeneous & Super Hard\\
    \hline
     {MMM3} & \tabincell{l}{1 Medivac, \\2 Marauders, \\8 Marines} & \tabincell{l}{1 Medivac, \\3 Marauders, \\9 Marines} & Asymmetric \& Heterogeneous & Super Hard\\
     \hline
     {MMM4} & \tabincell{l}{1 Medivac, \\3 Marauders, \\8 Marines} & \tabincell{l}{1 Medivac, \\4 Marauders, \\9 Marines} & Asymmetric \& Heterogeneous & Super Hard\\
    \hline
    {MMM5} & \tabincell{l}{1 Medivac, \\3 Marauders, \\8 Marines} & \tabincell{l}{1 Medivac, \\4 Marauders, \\10 Marines} & Asymmetric \& Heterogeneous & Super Hard\\
    \hline
    {MMM6} & \tabincell{l}{1 Medivac, \\3 Marauders, \\8 Marines} & \tabincell{l}{1 Medivac, \\4 Marauders, \\11 Marines} & Asymmetric \& Heterogeneous & Super Hard\\
    \hline
  \end{tabular}
  
\end{table*}

\begin{table*}[h]
  \caption{Description of sz\_series SMAC maps. }
  \label{SMACsz}
  \centering
  \newcommand{\tabincell}[2]{\begin{tabular}{@{}#1@{}}#2\end{tabular}}
  \begin{tabular}{|c|c|c|c|c|}
    \hline
    Map Name & Ally Units & Enemy Units & Type & Difficulty\\
     \hline
    {2s3z} & \tabincell{l}{2 Stalkers, \\3 Zealots} & \tabincell{l}{2 Stalkers, \\3 Zealots} & Symmetric \& Heterogeneous & Easy\\
     \hline
    2s3z\_vs\_2s4z & \tabincell{l}{2 Stalkers, \\3 Zealots} & \tabincell{l}{2 Stalkers, \\4 Zealots} & Symmetric \& Heterogeneous & Hard\\
    \hline
    3s4z & \tabincell{l}{3 Stalkers, \\5 Zealots} & \tabincell{l}{3 Stalkers, \\4 Zealots} & Symmetric \& Heterogeneous & Easy\\
     \hline
    3s5z & \tabincell{l}{3 Stalkers, \\5 Zealots} & \tabincell{l}{3 Stalkers, \\5 Zealots} & Symmetric \& Heterogeneous & Easy\\
         \hline
    3s5z\_vs\_3s6z & \tabincell{l}{3 Stalkers, \\5 Zealots} & \tabincell{l}{3 Stalkers, \\6 Zealots} & Symmetric \& Heterogeneous & Super Hard\\
         \hline
    {3s5z\_vs\_3s7z} & \tabincell{l}{3 Stalkers, \\5 Zealots} & \tabincell{l}{3 Stalkers, \\7 Zealots} & Symmetric \& Heterogeneous & Super Hard\\
    \hline
    4s7z & \tabincell{l}{4 Stalkers, \\7 Zealots} & \tabincell{l}{4 Stalkers, \\7 Zealots} & Symmetric \& Heterogeneous & Easy\\
     \hline
    {4s7z\_vs\_4s8z}& \tabincell{l}{4 Stalkers, \\7 Zealots} & \tabincell{l}{4 Stalkers, \\8 Zealots} & Symmetric \& Heterogeneous & Super Hard\\
    \hline
  \end{tabular}
  
\end{table*}

\begin{table*}[h]
  \caption{Description of m\_series SMAC maps. }
  \label{SMACm}
  \centering
  \newcommand{\tabincell}[2]{\begin{tabular}{@{}#1@{}}#2\end{tabular}}
  \begin{tabular}{|c|c|c|c|c|}
    \hline
    Map Name & Ally Units & Enemy Units & Type & Difficulty\\
        \hline
    3m\ & 3 Marines & 5 Marines & Symmetric \& Homogeneous & Easy\\      
    \hline
    4m\ & 4 Marines & 5 Marines & Symmetric \& Homogeneous & Easy\\
       \hline
    4m\_vs\_5m & 4 Marines & 5 Marines & Asymmetric \& Homogeneous & Hard\\
    \hline
    5m & 5 Marines & 5 Marines & Symmetric \& Homogeneous & Easy\\
        \hline
    5m\_vs\_6m & 5 Marines & 6 Marines & Asymmetric \& Homogeneous & Hard\\
        \hline
    6m & 6 Marines & 6 Marines & Symmetric \& Homogeneous & Easy\\
      \hline
    6m\_vs\_7m & 6 Marines & 7 Marines & Asymmetric \& Homogeneous & Hard\\
               \hline
    7m & 7 Marines & 7 Marines & Symmetric \& Homogeneous & Easy\\
        \hline
    7m\_vs\_8m & 7 Marines & 8 Marines & Asymmetric \& Homogeneous & Hard\\
                   \hline
    8m & 8 Marines & 8 Marines & Symmetric \& Homogeneous & Easy\\
            \hline
    8m\_vs\_9m & 8 Marines & 9 Marines & Asymmetric \& Homogeneous & Easy\\
     \hline
    9m & 9 Marines & 9  Marines & Symmetric \& Homogeneous & Easy\\            
    \hline
    9m\_vs\_10m & 9 Marines & 10 Marines & Asymmetric \& Homogeneous & Easy\\
     \hline
    10m & 10 Marines & 10 Marines & Symmetric \& Homogeneous & Easy\\   
       \hline
    10m\_vs\_11m & 10 Marines & 11 Marines & Asymmetric \& Homogeneous & Easy\\
    \hline
    10m\_vs\_12m & 10 Marines & 12 Marines & Asymmetric \& Homogeneous & Super Hard\\
    \hline
  \end{tabular}
  
\end{table*}

\subsection{Network Architecture and Hyperparameters}
\begin{table*}[h]
  \centering
  \caption{Hyperparameters concerning network structure in experiments.}
  \label{table:hyperparameters}
  \begin{tabular}{c|c}
        \toprule
        name & value \\
        \midrule
        mixing\_embed\_dim & 32 \\
        hypernet\_layers & 2 \\
        hypernet\_embed & 64 \\
        id\_length & 4 \\
        task\_repre\_dim & 32 \\
        state\_latent\_dim & 32 \\
        entity\_embed\_dim & 64 \\
        attn\_embed\_dim & 8 \\
        \bottomrule
  \end{tabular}
\end{table*}
Our implementation of Mattar is based on PyMARL\footnote{\url{https://github.com/oxwhirl/pymarl}} \cite{samvelyan2019starcraft} with StarCraft 2.4.6.2.69232 and uses its default hyperparameter settings. We apply the default $\epsilon$-greedy action selection algorithm to every algorithm, as $\epsilon$ decays from $1$ to $0.05$ in $50$K timesteps. We also adopt typical Q-learning training tricks like target networks and double Q-learning. Mattar has additional hyperparameters $\lambda_1, \lambda_2$, and $\lambda$ for doing representation learning, the scaling factors for observation prediction loss, reward prediction loss, and an entropy regularization term, respectively. We set these additional parameters to $1, 10$, and $0.1$ across all experiments. We use the default configurations for QMIX in the PyMARL framework. For RODE \cite{wang2021rode}, ASN~\cite{wang2019action},  QPLEX \cite{wang2020qplex}, QMIX~\cite{rashid2018qmix}, and UPDET~\cite{hu2021updet}, we use the codes provided by the authors from their original papers with default hyperparameters settings. For the hyperparameters concerning network structure, our selection is listed in Table~\ref{table:hyperparameters}. We used this set of hyperparameters in all experiments.

\subsection{Experimental Details}\label{appx:experimental details}
Our experiments were performed on a desktop machine with 2 NVIDIA GTX 2080 Ti GPUs. For all the performance curves in our paper, we pause training every $10$K timesteps and evaluate for $32$ episodes with decentralized greedy action selection. We evaluate the test win rate, the percentage of episodes in which the agents defeat all enemies within the time limit in $32$ testing episodes for SMAC. For each part of experiments in our paper, descriptions about experimental details are as follows:

\textbf{Generalizability to unseen tasks} \ \ For each compared algorithm, we carried out $5$ experiments with different random seeds. In each experiment, we evaluate trained model for $32$ episodes on target task. The results recorded in Tables~\ref{table:sz test}$\sim$\ref{table:m test} are the mean results for these $5$ random seeds.

\textbf{Task representations provide a good initialization for fine-tune} \ \ For the performance of transfer learning, we trained $2$ source models with different random seeds for each map and carried out transfer learning experiments with $2$ random seeds for each source model. For the performance of learning from scratch, we carried out $4$ experiments with different random seeds for each map.

\textbf{Benefits of multi-task learning} \ \ We carried out $5$ experiments with different random seeds for both multi-task learning and learning on single task. For the experiments of multi-task learning on three tasks shown in the paper, the sets of tasks are \{5m, 5m\_vs\_6m, 8m\_vs\_9m, 10m\_vs\_11m\}, \{2s3z, 3s5z, 3s5z\_vs\_3s6z\}, and \{MMM, MMM2, MMM4\}, respectively.

\textbf{Bonus: performance on single-task training} \ \ For this experiment, we carried out $5$ experiments for each compared algorithm, and did evaluation during training process as we described above.

\subsection{Forward Model for Task Representation Learning}\label{appx:architecture}
\begin{figure*}[t!]
    \centering
    \vspace{-1em}
    \subfigure[Network architecture for forward model\label{fig:fm}]{\includegraphics[width=0.45\textwidth]{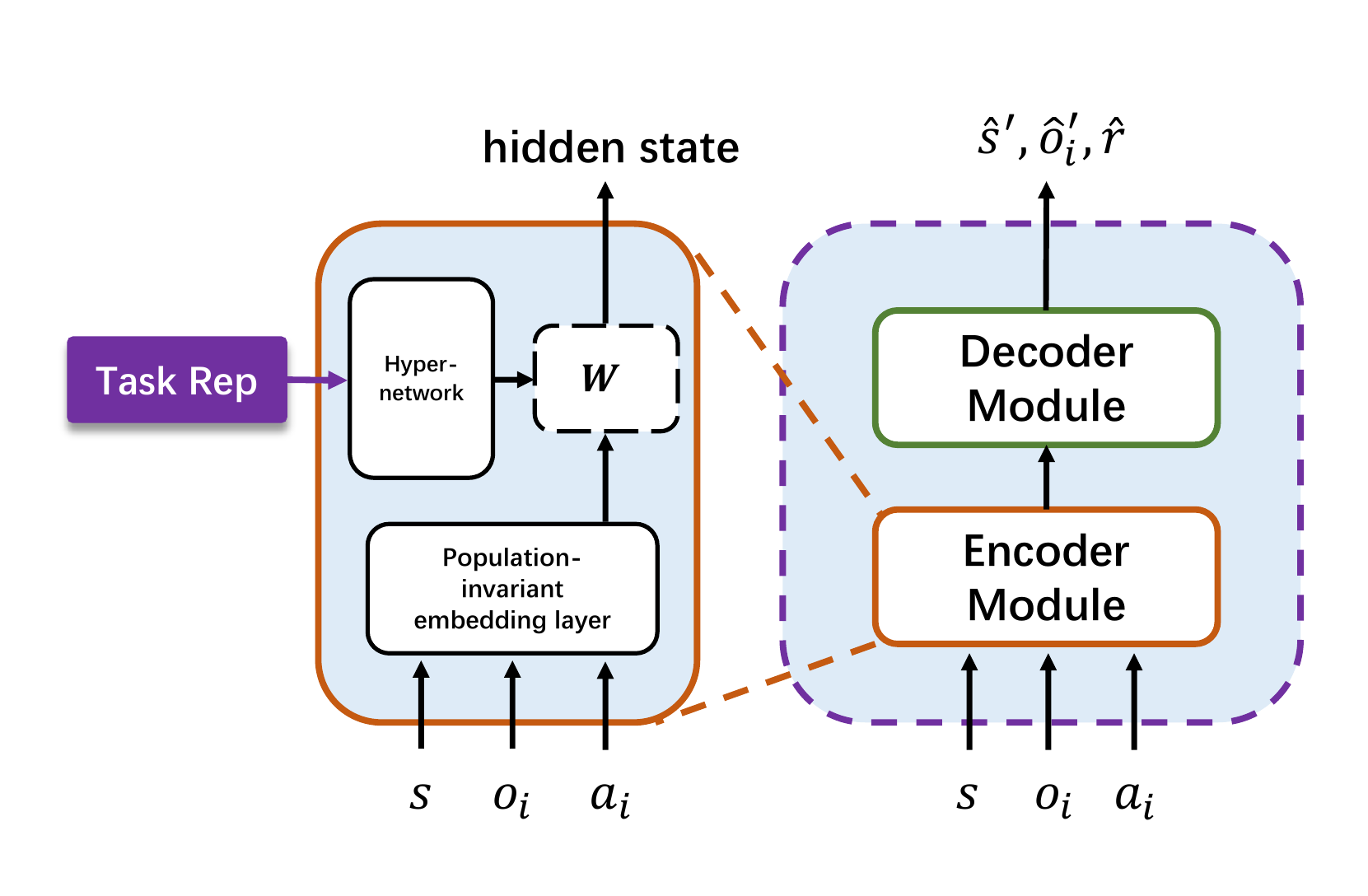}}
    \subfigure[Network architecture for interaction actions\label{fig:dan}]{\includegraphics[width=0.3\textwidth]{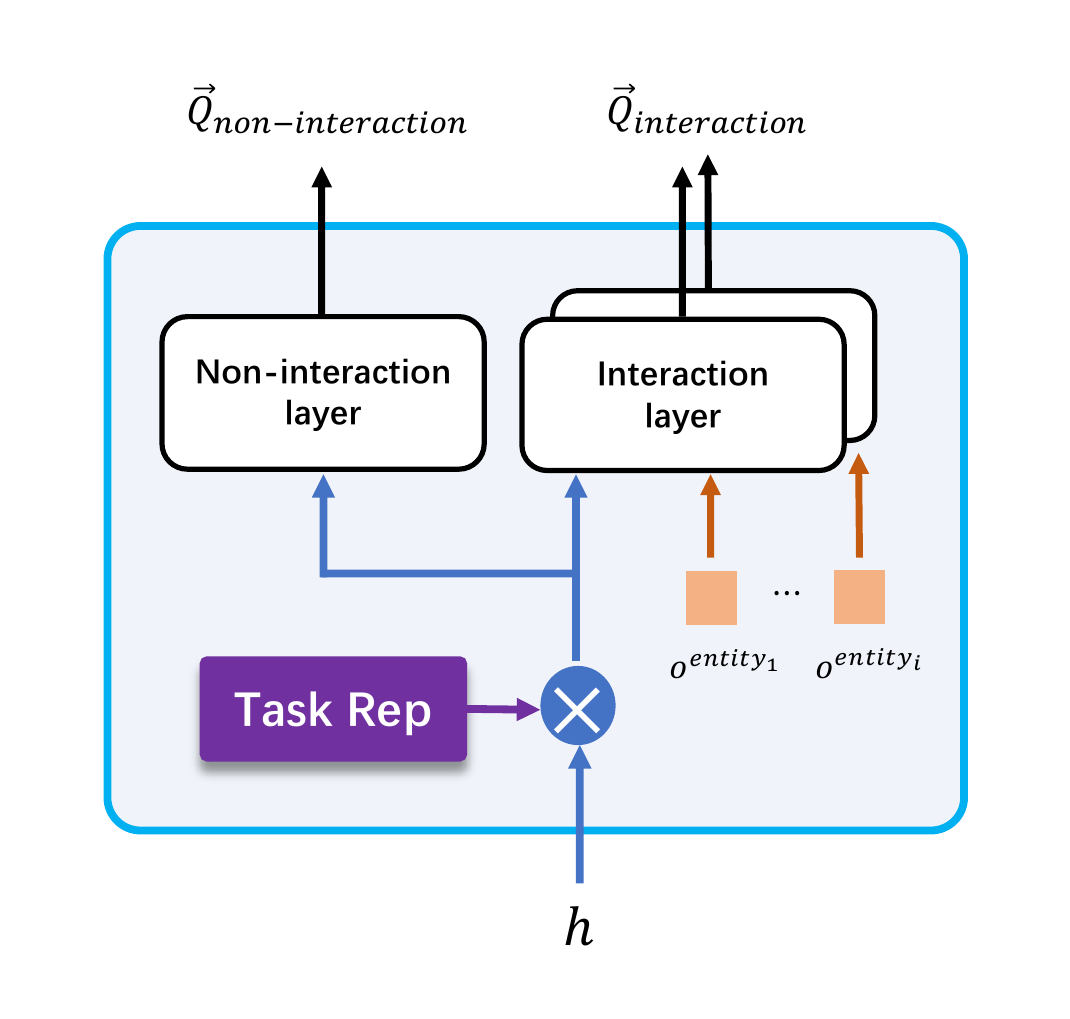}}
    \caption{Supplemental descriptions about network
    architecture for forward model and interaction actions.}
    \vspace{-1em}
\end{figure*}

In our method, we utilize forward model learning to help build task representations which can capture the similarity between different tasks. We use a hypernetwork as representation explainer to generate the parameters of forward model. In practical implementation, we design the forward model as two components, an encoder and a decoder (Fig.~\ref{fig:fm}). We use similar techniques to those used in designing Q-value functions to make the encoder a population-invariant structure and let the decoder be a task-specific structure.

For the encoder network, we first use the population-invariant embedding layer to get a fixed-dimension embedding vector and feed it into a fully-connected layer generated by the hypernetwork representation explainer. The output hidden variable is then fed into the decoder to predict the next state, the next observation, and the global reward. The encoder module and the hypernetwork are shared among tasks and are fixed when learning new task representations, while the decoder module is task-specific, and we allow the decoder to be optimized together with task representations when adapting to unseen tasks. This solution is a trade-off between allowing a designed forward model expressive enough to solve the forward-prediction problems in different tasks by using individual decoders and capturing the similarity between tasks by sharing the hypernetwork and the encoder module which are core components of the forward model.

For the population-invariant structure in the encoder module, we decompose the input state and observation into several entity-specific components, pass them through an embedding layer, respectively, and do pooling operation for their output vectors. We also decompose the action input to deal with the situation of dynamic-dimension action input. We incorporate the decomposed action to observation $o_i$, concatenating non-interaction portions with agent $i$'s own observation component $o^{own}_i$ and interaction portions with corresponding entity's observation component. We claim that other population-invariant structures can be applied to our approach to polish our work further.

\subsection{Performance of Mattar on More SMAC Maps for Single-Task Learning}\label{appx:full performance}
\begin{figure*}[t!]
\centering
\subfigure[1c3s5z]{\includegraphics[width=0.3\linewidth]{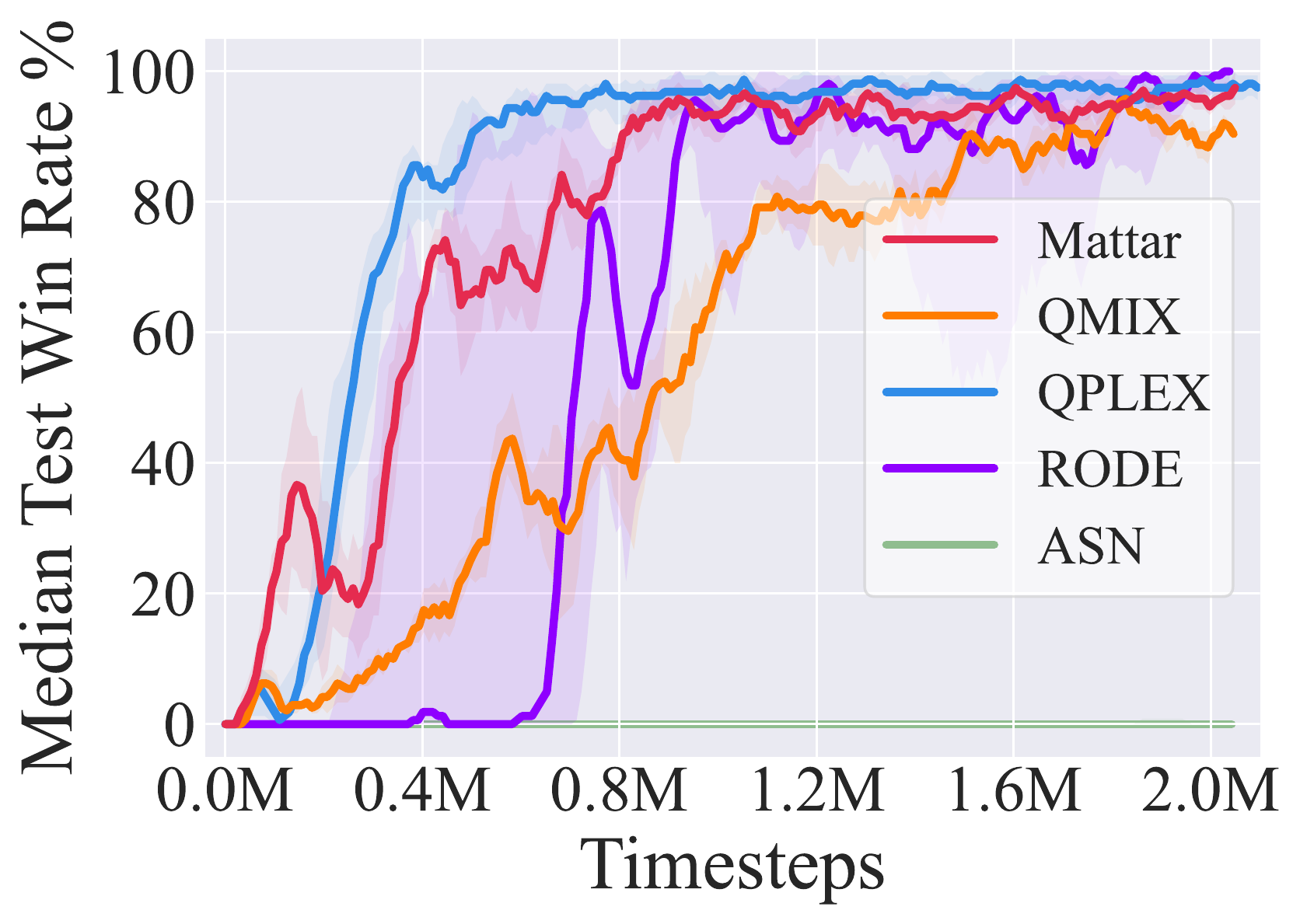}}%
\subfigure[2c\_vs\_64zg]{\includegraphics[width=0.3\linewidth]{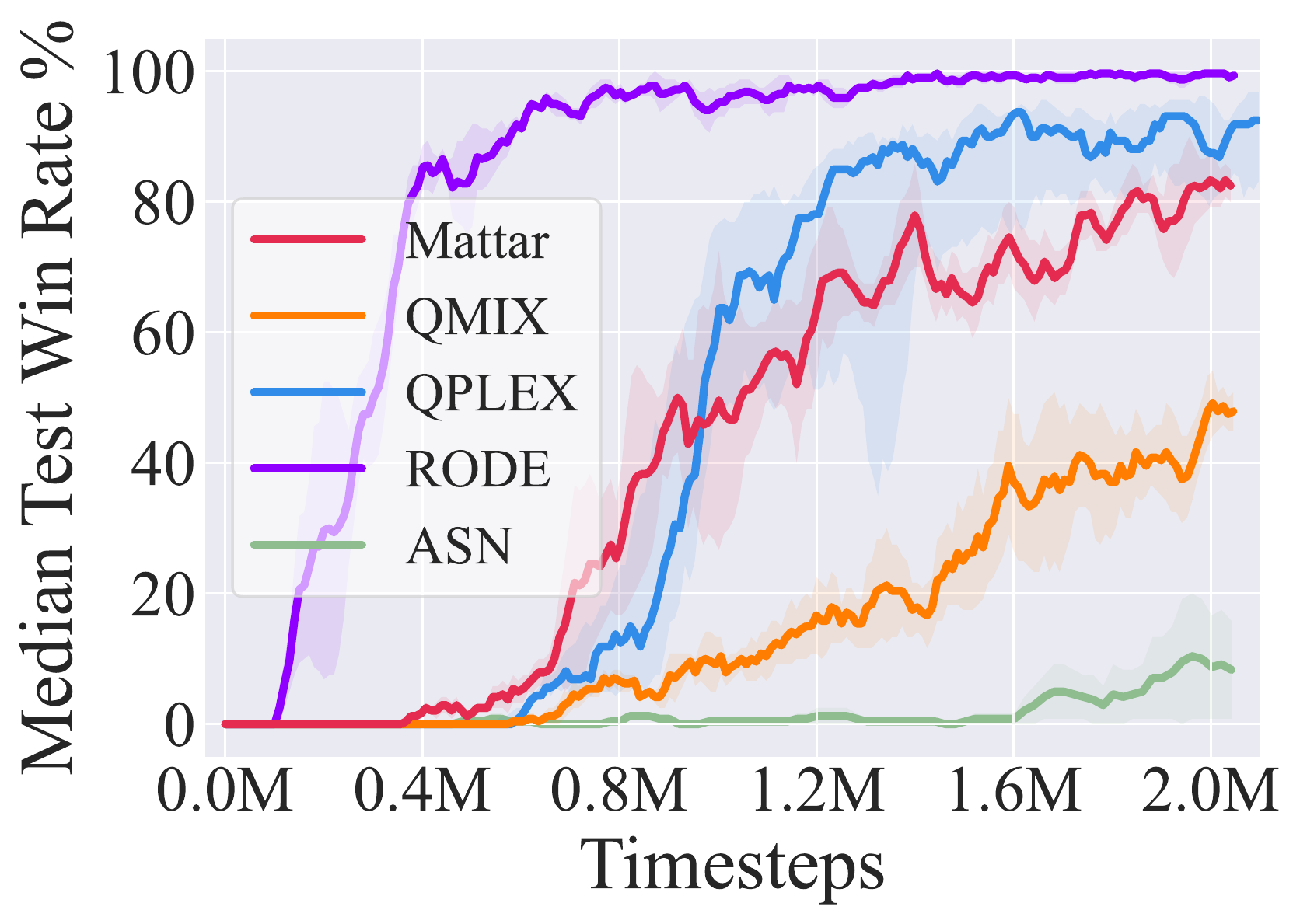}}%
\subfigure[2s\_vs\_1sc]{\includegraphics[width=0.3\linewidth]{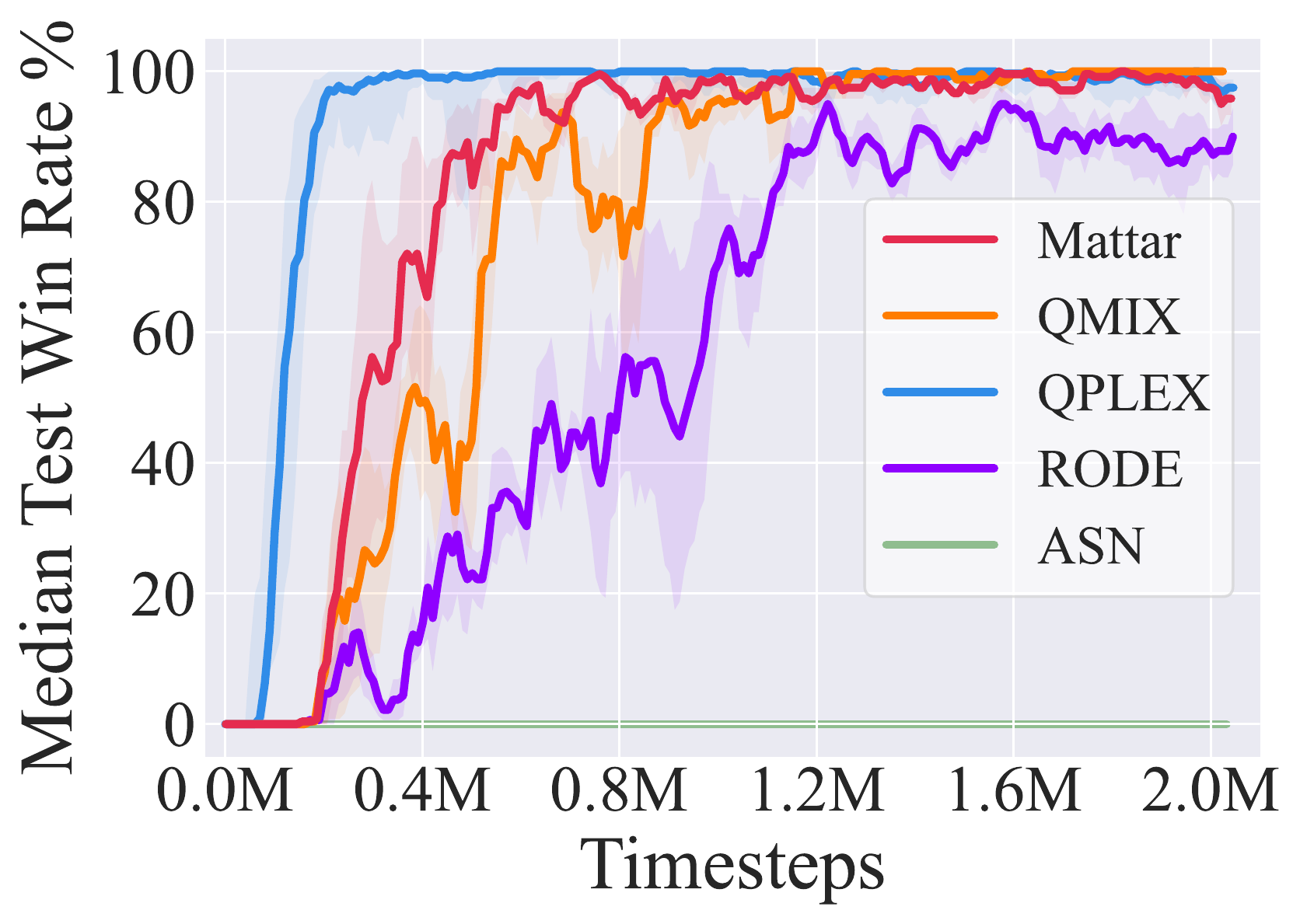}}%
\qquad
\vspace{-1em}
\subfigure[2s3z]{\includegraphics[width=0.3\linewidth]{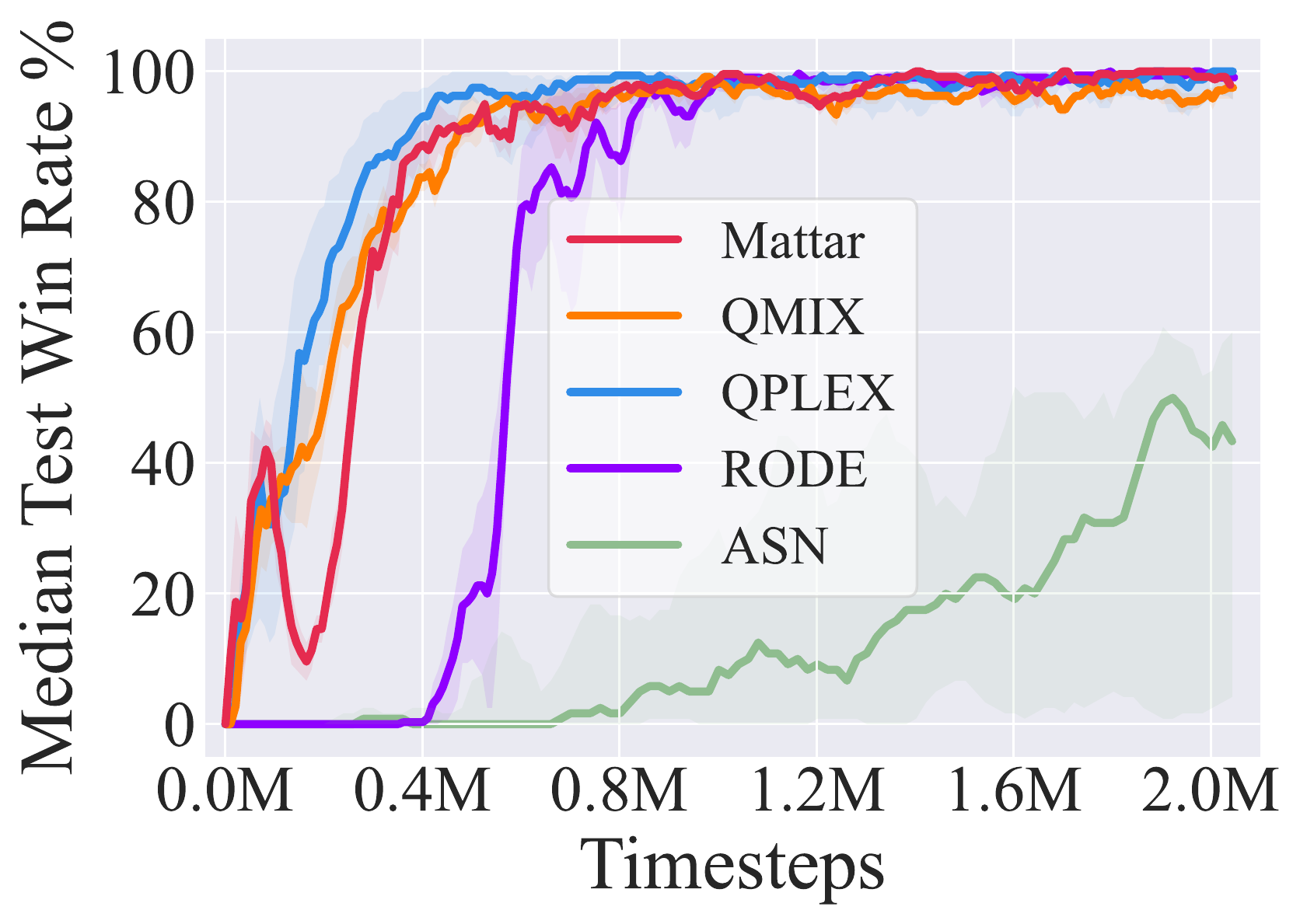}}%
\subfigure[3s\_vs\_5z]{\includegraphics[width=0.3\linewidth]{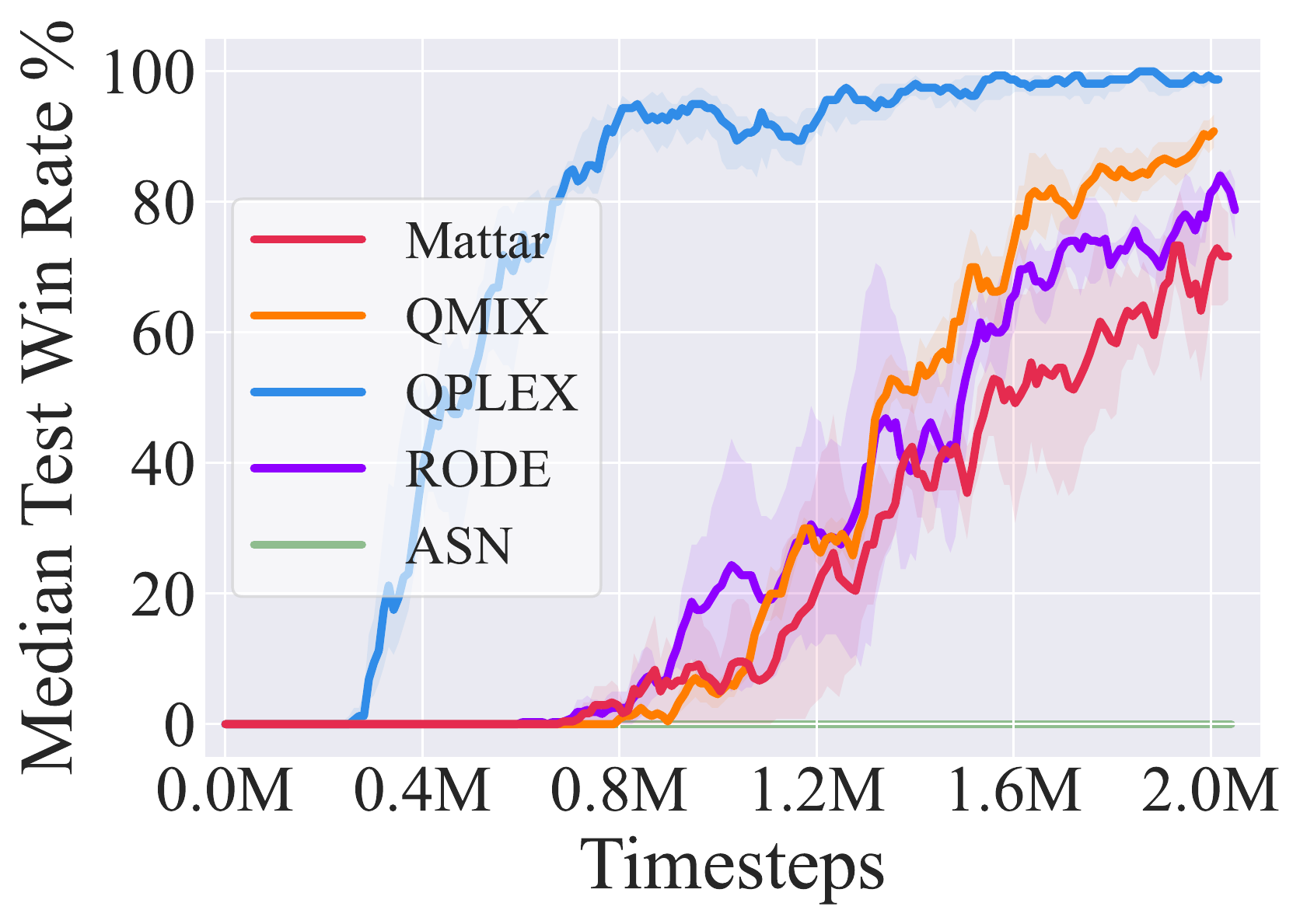}}%
\subfigure[3s5z\_vs\_3s6z]{\includegraphics[width=0.3\linewidth]{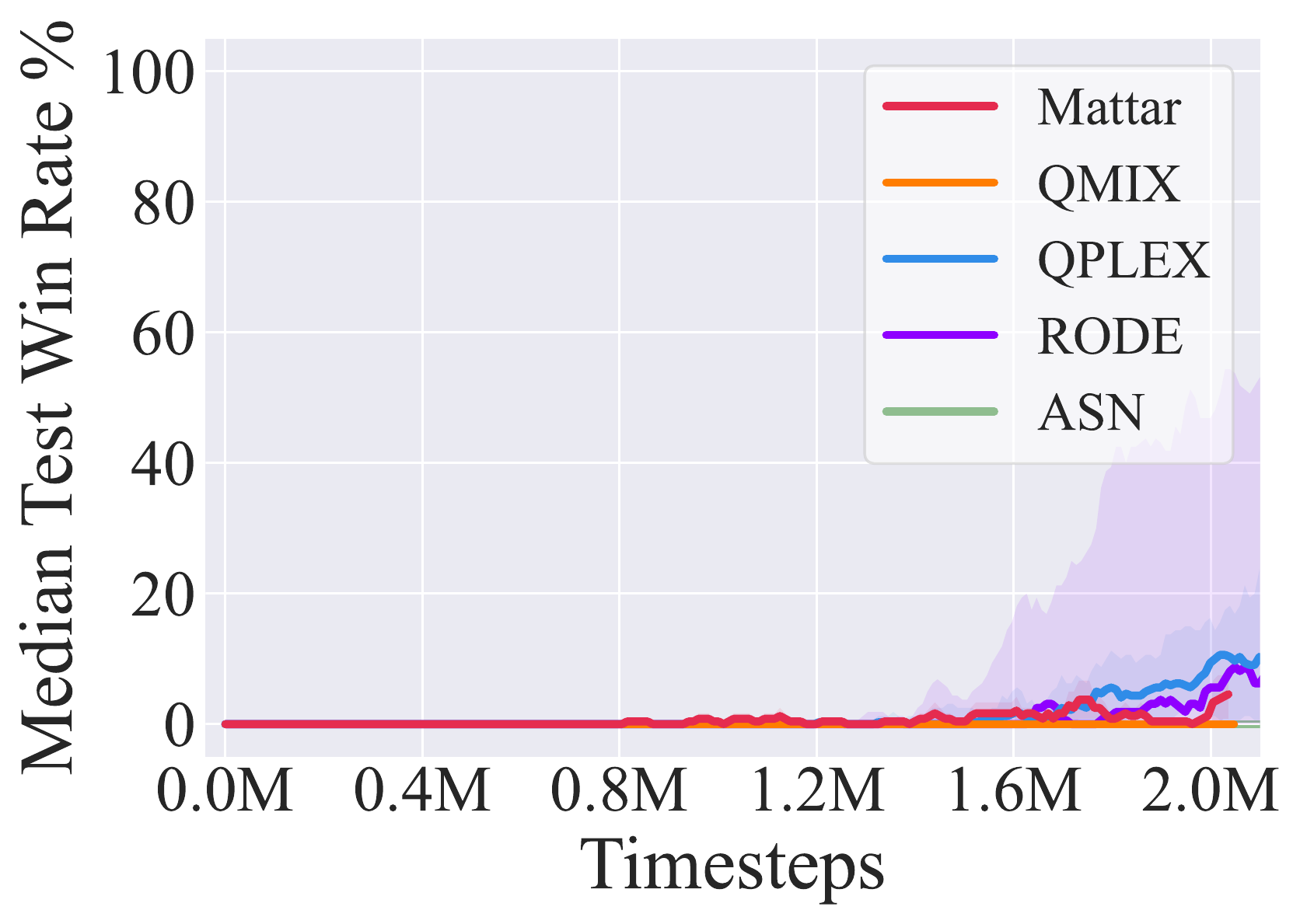}}%
\qquad
\vspace{-1em}
\subfigure[6h\_vs\_8z]{\includegraphics[width=0.3\linewidth]{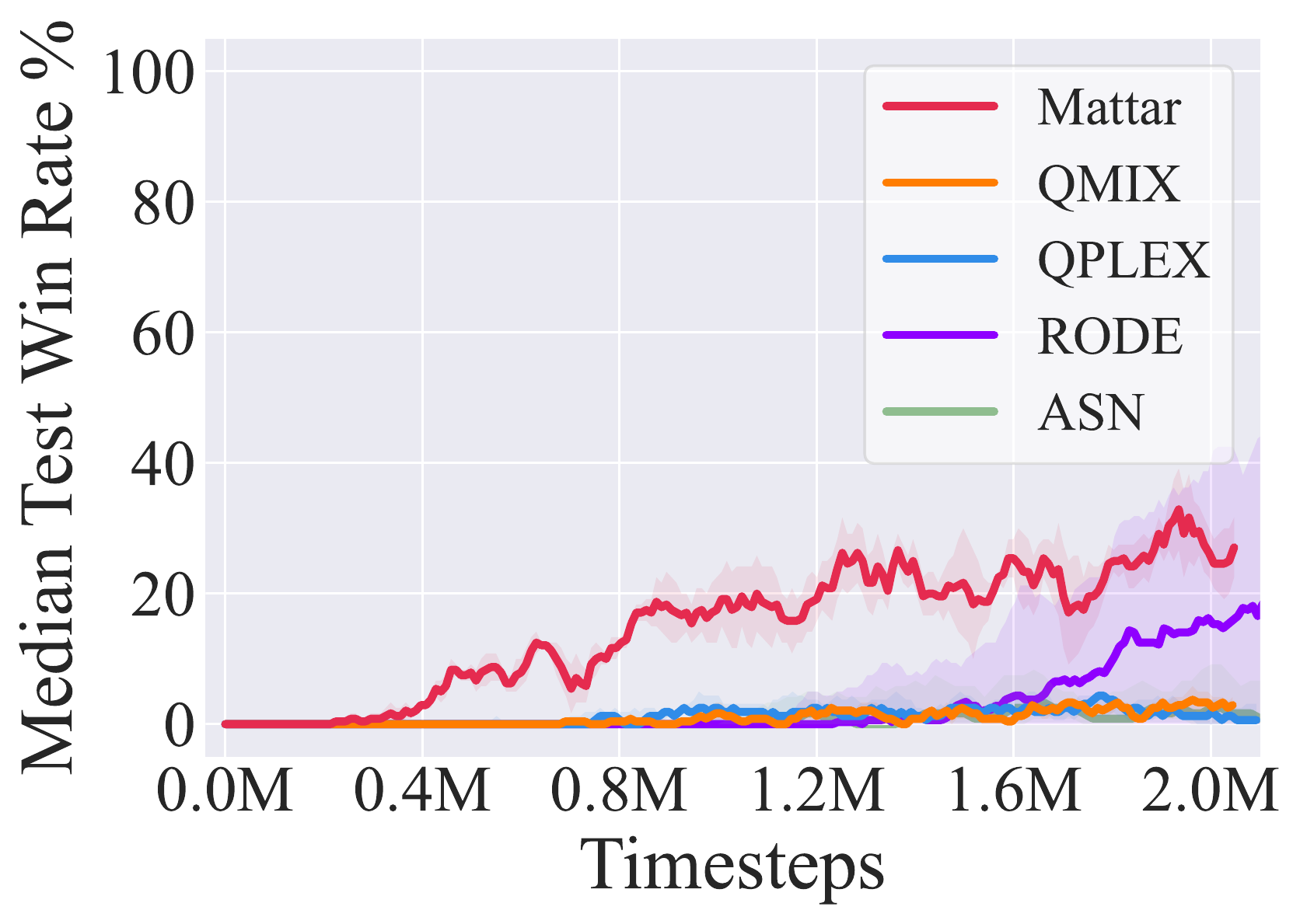}}%
\subfigure[10m\_vs\_11m]{\includegraphics[width=0.3\linewidth]{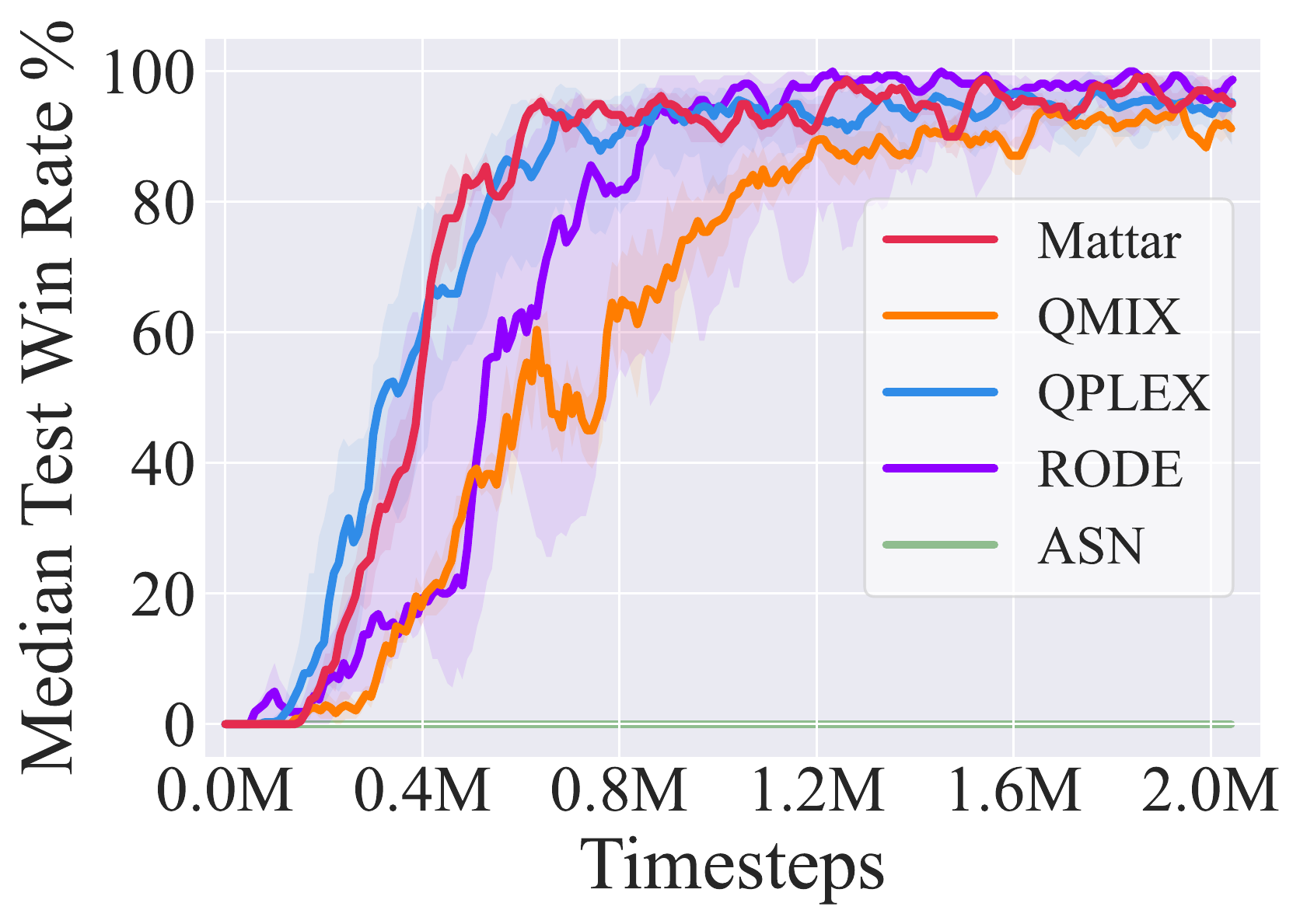}}%
\subfigure[corridor]{\includegraphics[width=0.3\linewidth]{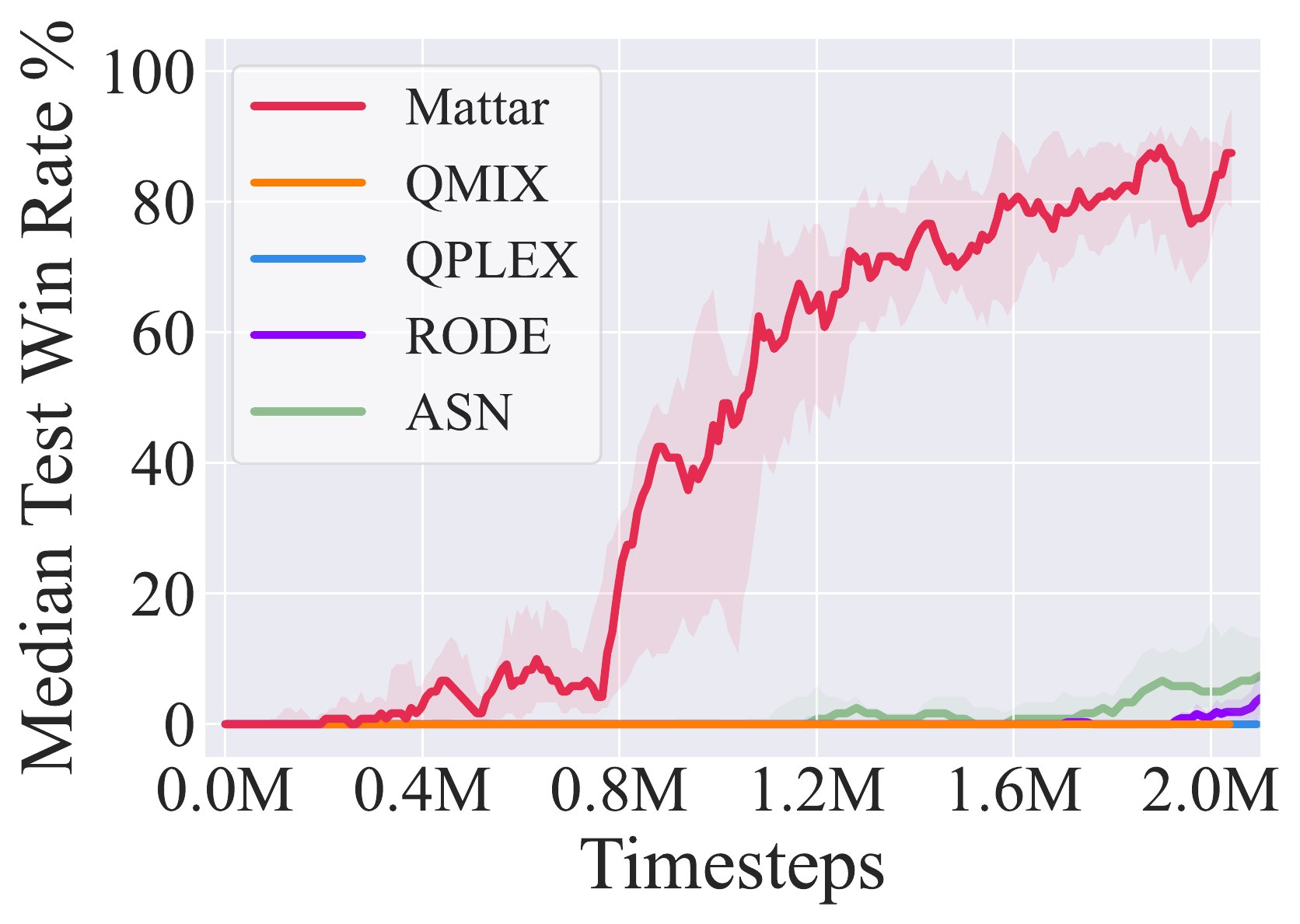}}%
\centering
\caption{Supplemental results for the performance of \name~on SMAC benchmark when learning from scratch on single tasks.}
\label{fig:full performance}
\vspace{-1em}
\end{figure*}

The additional results are shown in Fig.~\ref{fig:full performance}. 
\subsection{Dynamic Action Dimensions}\label{appx:interaction action}
In some multi-agent environments, there exist interaction actions that have semantics relating to an entity. In this case, the action dimension is related to the number of agents in the environment. For this problem, we design a particular structure to calculate Q-values for these interaction actions, as shown in Fig.~\ref{fig:dan}.


With the help of previous techniques, we have already obtained a fixed-dimension embedding vector $h$ for dynamic-dimension observation. For non-interaction actions, we use a fully-connected network to compute the Q-value vector with the concatenation of $h$ and task representation as input. For an interaction action, we use an action-sharing network, which takes as input the concatenation of $h$, task representation $z$, and observation component relating to the entity for that interaction action, to output a scalar as the Q-value of that action.


\end{document}